\PassOptionsToPackage{table}{xcolor}
\documentclass[sigconf,10pt]{acmart}
\makeatletter
\let\@ACM@checkaffil\@empty
\makeatother
\usepackage{xcolor}
\usepackage{comment}
\usepackage{xspace}
\usepackage{makecell}
\usepackage{algpseudocode}
\usepackage{subfig}
\usepackage{booktabs}
\usepackage{array}
\usepackage{caption}
\usepackage[export]{adjustbox}
\usepackage{multirow}
\usepackage{color}
\usepackage{bm}

\usepackage{amsmath}
\usepackage{amssymb}
\usepackage{wasysym}
\usepackage{pifont}
\usepackage{enumitem}
\usepackage{soul}
\usepackage{tikz}
\usepackage[percent]{overpic}
\usepackage[export]{adjustbox}
\newcolumntype{L}[1]{>{\raggedright\arraybackslash}p{#1}}
\newcolumntype{C}[1]{>{\centering\arraybackslash}p{#1}}
\allowdisplaybreaks

\newcommand\revision[1]{\textcolor{black}{#1}}

\copyrightyear{2026}
\acmYear{2026}
\setcopyright{cc}
\setcctype{by}
\acmConference[MobiCom '26]{The 32nd Annual International Conference on Mobile Computing and Networking}{October 26--30, 2026}{Austin, TX, USA}
\acmBooktitle{The 32nd Annual International Conference on Mobile Computing and Networking (MobiCom '26), October 26--30, 2026, Austin, TX, USA}
\acmDOI{10.1145/3795866.3844478}
\acmISBN{979-8-4007-2505-0/2026/10}

\newlength{\mcolw}

\begin{document}
\settopmatter{printacmref=true}

\newcommand{\name}{\textit{GRADE}\xspace}

\title[GRADE: Single-Frame Radar Depth Estimation Under Visual Degradation]{\name: Single-Frame Generative Radar Depth Estimation Under Visual Degradation}

\author{Bin Zhao}
\orcid{0009-0006-6029-6511}
\email{bz35@rice.edu}
\affiliation{%
  \institution{Rice University}
}

\author{Patrick Chiou}
\orcid{0009-0004-2519-3854}
\email{pc82@rice.edu}
\affiliation{%
  \institution{Rice University}
}

\author{Nakul Garg}
\orcid{0000-0002-8585-0180}
\email{nakul@rice.edu}
\affiliation{%
  \institution{Rice University}
}

\begin{abstract}
Dense 3D depth perception fails under smoke, fog, and darkness because optical sensors cannot penetrate airborne particulates. 
mmWave radar remains usable and measures range accurately under these conditions, but its small aperture limits angular resolution. We present \textit{GRADE}, which grounds a pretrained generative prior in single-frame radar geometry to estimate high-fidelity metric depth. \textit{GRADE} first maps raw 4D radar spectra to coarse metric depth. A latent diffusion backbone then recovers structural detail while conditioning every denoising step on this estimate. A pixel-space adapter uses residual camera cues when available and \revision{is trained across clear, smoke-degraded, and occluded inputs so the full output approaches the radar-conditioned path as visibility degrades.} \revision{Trained and evaluated} on ${\sim}$95K frames across 12 buildings with real smoke, \textit{GRADE} achieves an MAE of \revision{0.303\,m} in clear scenes and \revision{0.313\,m} under smoke, outperforming existing baselines.
Code and datasets are available at \textcolor{blue}{\url{https://phi-lab-rice.github.io/GRADE}}.
\end{abstract}



\maketitle
\section{Introduction}
\label{sec:intro}

\noindent\textbf{Motivation.}
Robotic navigation, AR-guided assembly, and search-and-rescue depend on real-time 3D reconstruction of the surrounding geometry~\cite{intro-1-1,intro-1-2,intro-1-3,intro-1-4,intro-1-5}. Cameras and LiDAR provide detailed depth under clear conditions ~\cite{da3,marigold}, but smoke and fog attenuate visible and near-infrared light, and passive cameras require ambient illumination~\cite{intro-2-1,intro-2-2,intro-2-3}. A first responder in smoke or a robot in dust can therefore lose optical depth when it is needed most.
mmWave radar provides a complementary measurement. At 77\,GHz, it requires no illumination and remains usable through smoke, fog, and dust~\cite{guan2020through,prabhakara2023radarhd}. Single-chip FMCW hardware also supports compact mobile platforms~\cite{intro-3-1}. These properties make radar \revision{the natural candidate for the geometric backbone of} depth perception under visual degradation. The remaining challenge is fidelity: can a small radar aperture recover the detail these applications require?

\noindent\textbf{The resolution bottleneck.}
The angular resolution of a mmWave radar is governed by its aperture, and a commercial single-chip device with a small virtual antenna array produces depth estimates that are angularly coarse and extremely sparse~\cite{grt,panoradar}. This is why walls may appear as blurred regions, furniture edges are usually absent, and small objects can be missed entirely~\cite{prabhakara2023radarhd}. 
The output captures the metric \emph{layout} of a scene but its geometric \emph{structure} is lost. This gap limits AR overlays, obstacle boundary delineation, and 3D mapping.

\noindent\textbf{\revision{Why single-frame?}}
\revision{Multi-frame fusion and SAR improve angular resolution when controlled motion and accurate pose are available. These assumptions can fail under irregular wearable motion, on small drones without reliable odometry, and in stop-and-go emergency response, where geometry is needed before a scan completes. \name{} targets this per-frame regime and remains complementary to SAR when its acquisition assumptions hold.}

\begin{figure*}[t]
    \vspace{-0.1in}
    \centering
    \setlength{\tabcolsep}{0pt}
    \renewcommand{\arraystretch}{1.0}

    \newlength{\teaserimgw}
    \setlength{\teaserimgw}{0.138\textwidth} %

    \newcommand{\teaserimg}[1]{%
    \begin{overpic}[width=\teaserimgw,valign=t]{#1}
    \end{overpic}%
    }

    \newcommand{\teaserimglabel}[2]{%
    \begin{overpic}[width=\teaserimgw,valign=t]{#1}
        \put(4,40){%
        \colorbox{black!55}{%
        \textcolor{white}{\tiny\bfseries\strut #2}%
        }%
        }
    \end{overpic}%
    }

    \newcommand{\teaserlabel}[1]{%
    \parbox[c]{\teaserimgw}{\centering\small\bfseries #1}%
    }

    \begin{tabular}{@{}ccccccc@{}}

        \teaserimglabel{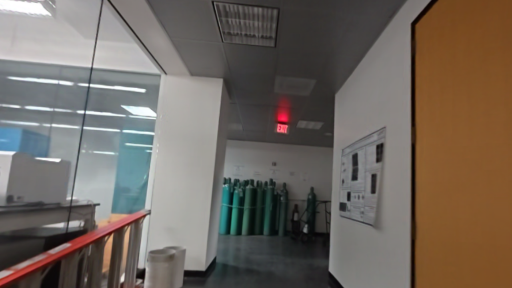}{Clear} &
        \teaserimg{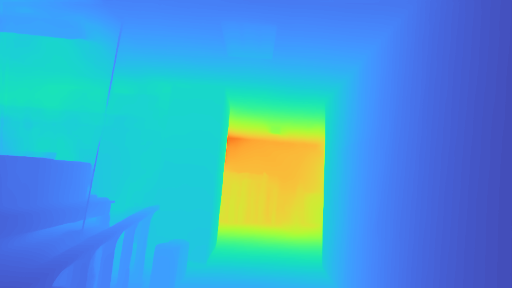} &
        \teaserimg{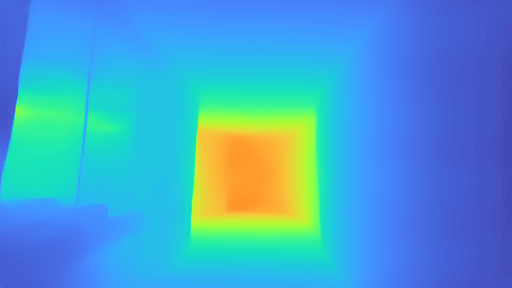} &
        \teaserimg{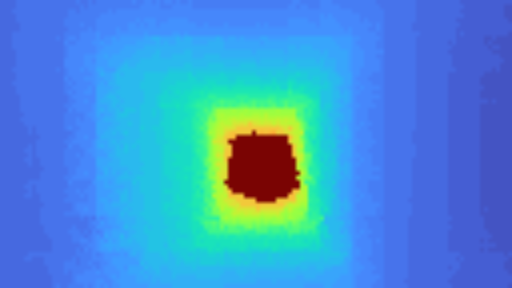} &
        \teaserimg{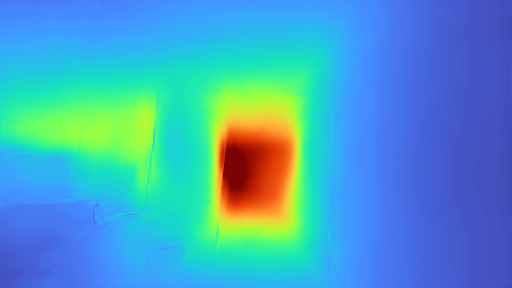} &
        \teaserimg{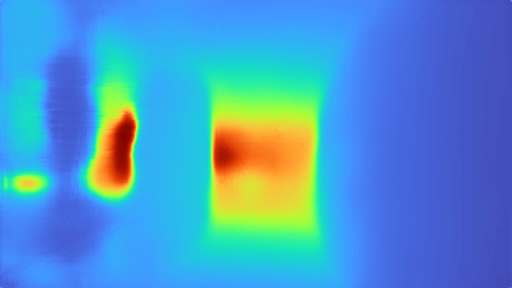} &
        \teaserimg{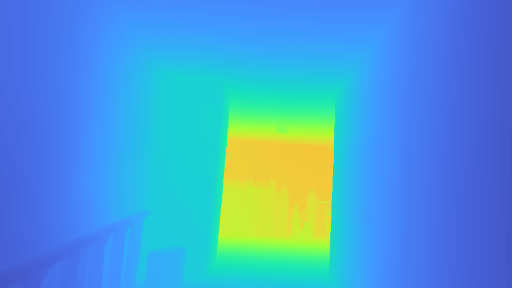} \\[-4pt]

        \teaserimglabel{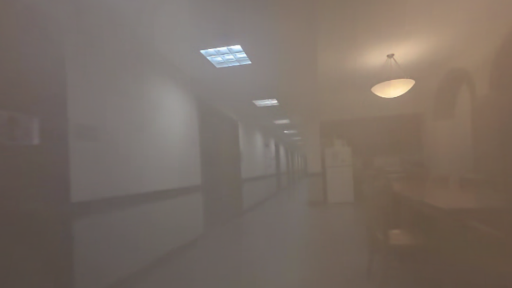}{Medium Smoke} &
        \teaserimg{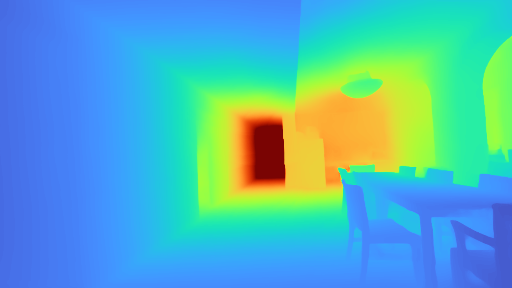} &
        \teaserimg{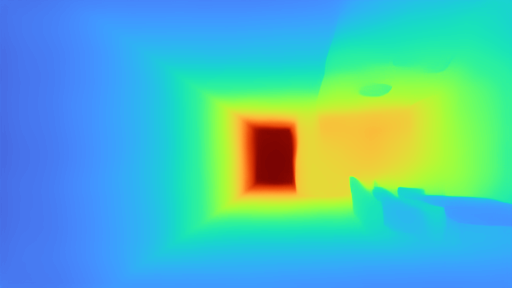} &
        \teaserimg{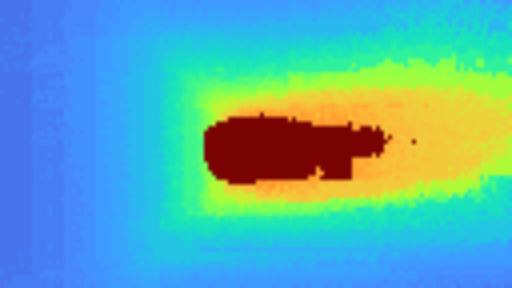} &
        \teaserimg{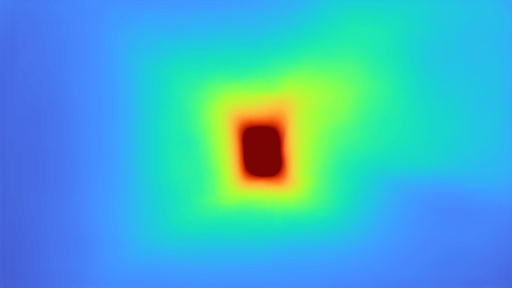} &
        \teaserimg{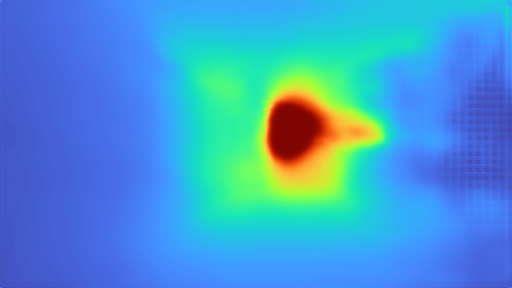} &
        \teaserimg{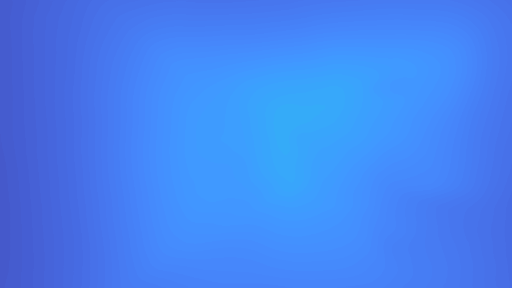} \\[-4pt]

        \teaserimglabel{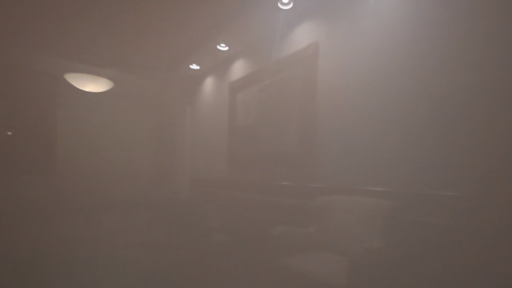}{Heavy Smoke} &
        \teaserimg{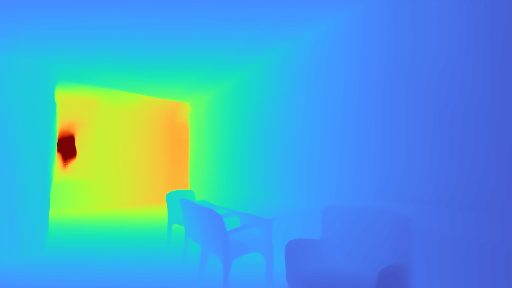} &
        \teaserimg{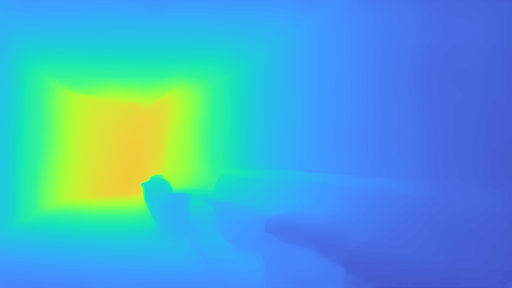} &
        \teaserimg{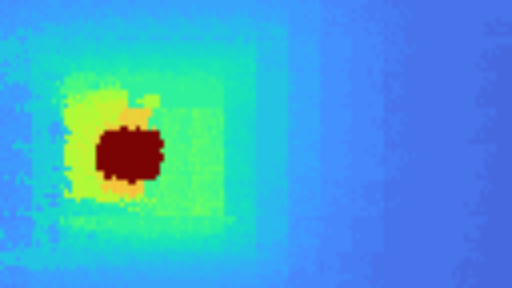} &
        \teaserimg{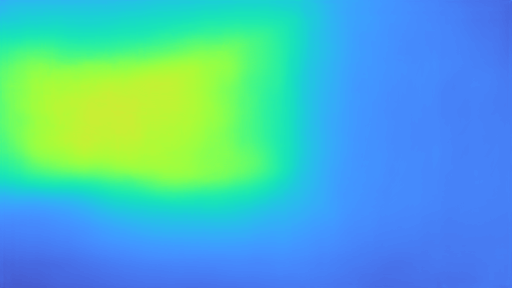} &
        \teaserimg{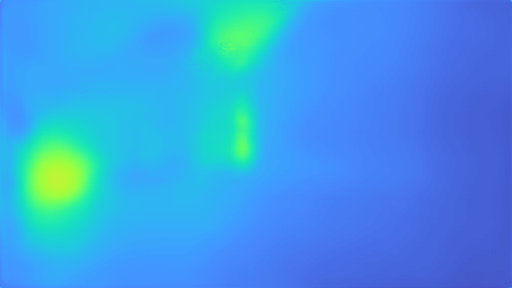} &
        \teaserimg{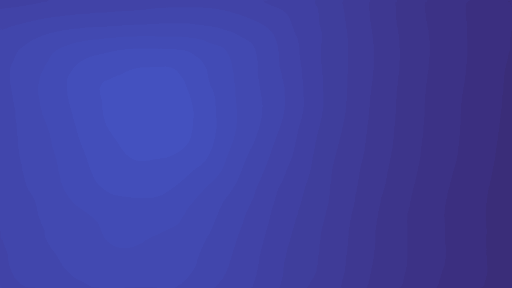} \\[-2pt]

        \teaserlabel{Input RGB} &
        \teaserlabel{Ground Truth} &
        \teaserlabel{\textit{\name} (Ours)} &
        \teaserlabel{GRT~\cite{grt}} &
        \teaserlabel{\revision{RadarCam-Depth~\cite{li2024radarcam}}} &
        \teaserlabel{CaFNet~\cite{sun2024cafnet}} &
        \teaserlabel{DA3~\cite{da3}} \\

    \end{tabular}
        \caption{\revision{\small{\name{} estimates depth from clear through heavy-smoke conditions while preserving more scene structure than the evaluated baselines.}}}
    \label{fig-teaser}
\end{figure*}

\noindent\textbf{Prior approaches and their assumptions.}
Prior work falls into three paradigms, each limited by a core assumption. \emph{Multi-modal fusion methods}~\cite{sun2024cafnet,singh2023depth,wang2025tacodepth,huang2025l4dr,palladin2024samfusion} compensate for radar sparsity by fusing it with a dense camera stream, treating the camera as the primary feature source. When the camera degrades, the backbone collapses: Depth Anything 3~\cite{da3} sees its Chamfer Distance rise from 0.196\,m$^2$ to 4.407\,m$^2$ under smoke ($>$22$\times$). \emph{SAR systems}~\cite{panoradar,saadat2020millicam,dodds2025mito,gao2021mimo} achieve high resolution by synthesizing large virtual apertures through controlled motion. PanoRadar~\cite{panoradar} forms a 9{,}600-element cylindrical array via rotation, achieving LiDAR-comparable imaging, but SAR requires sub-wavelength positioning accuracy ($\lambda/2 = 1.9$\,mm at 79\,GHz) and sufficient spatial displacement, \revision{which delays immediate depth on static or stop-and-go platforms and becomes difficult under irregular motion or unreliable odometry.} \emph{Single-frame radar methods} ~\cite{grt,prabhakara2023radarhd,zhou2025rise} eliminate both optical dependence and motion requirements. GRT~\cite{grt} predicts 3D occupancy from a single frame, but its output remains bounded by the physical aperture (LPIPS above 0.44). \revision{To our knowledge, prior single-frame radar systems have not combined egocentric 3D depth, generative priors and degraded-vision operation.}

\noindent\textbf{Leveraging generative priors.}
A compelling opportunity to close this gap without enlarging the aperture comes from vision foundation models. Diffusion models pretrained on web-scale image and depth data encode rich structural priors about how surfaces terminate, edges connect, and objects relate spatially~\cite{ldm,marigold,lotus}. Marigold~\cite{marigold} and Lotus~\cite{lotus} showed that repurposing these priors for monocular depth yields state-of-the-art perceptual quality. In this paper we ask whether these learned world priors can be grounded in the metric geometry that radar measures, producing high-fidelity \emph{3D} depth from a \textit{single radar frame without SAR, without motion, and without a reliable camera}. This reframing, from ``improving radar resolution'' to ``anchoring pretrained generative knowledge in single-frame radar geometry,'' is the central idea of this paper.

\noindent\textbf{Technical challenges.}
Realizing this idea is nontrivial. Three fundamental challenges must be solved. 
\textbf{(C1) The radar-to-vision domain gap.} Raw 4D radar spectra are complex-valued RF tensors entirely outside the training distribution of any vision foundation model. Applying a pretrained diffusion backbone directly to radar data produces incoherent outputs; the signal must be explicitly translated into a vision-compatible depth representation first. \textbf{(C2) Hallucination under geometric ambiguity.} \revision{Single-frame sparse-to-dense reconstruction is one-to-many. When the conditioning signal is ambiguous, the diffusion prior can generate a visually plausible but inaccurate depth map.} For instance, in our experiments (see Fig.~\ref{fig:diffusion_hallucination}) the model correctly identifies a staircase-like structure but places it at the wrong location. The prior knows \emph{what} belongs indoors but not \emph{where} in this specific scene.
\textbf{(C3) Leveraging degraded visual cues without dependence.} A degraded RGB image often retains edges and contrasts sufficient to suppress hallucination, \revision{yet naive fusion~\cite{sun2024cafnet,wang2025tacodepth,singh2023depth} that treats RGB as a symmetric input collapses when the camera fails. CaFNet's CD increases from 0.174 to 1.846\,m$^2$ from clear to smoke in our evaluation~\cite{sun2024cafnet}. Our goal is to use visual cues when available while retaining the radar-conditioned prediction when the image is unusable.}

\noindent\textbf{Our approach.}
We present \name{}\footnote{\name{} stands for \textbf{G}enerative \textbf{RA}dar \textbf{D}epth \textbf{E}stimation}, a two-stage framework that grounds pretrained generative priors in single-frame radar geometry for high-fidelity metric depth under visual degradation. In Stage~1, a Radar Depth Module translates the raw 4D radar spectrum into a coarse, metrically grounded depth image via a transformer encoder-decoder, bridging the RF-to-vision domain gap (\S\ref{sec:radar_module}). In Stage~2, a Diffusion Depth Refinement Module conditions a pretrained latent diffusion backbone on this radar depth, recovering structural detail through learned world priors while remaining anchored to the radar's metric geometry at every denoising step (\S\ref{sec:diffusion_module}). To suppress hallucination, an RGB Visual Guidance Module operates in parallel: a ControlNet-style adapter extracts residual spatial cues from the camera in pixel space and injects them via zero-initialized skip connections (\S\ref{sec:rgb_module}). 
\revision{Training across clear, smoke-degraded, and fully occluded inputs exposes the adapter to different levels of visual evidence. As visibility degrades, the full model's output approaches the radar-conditioned prediction; \S\ref{sec:graceful} quantifies where visual guidance helps.}
\newline

\noindent\textbf{Summary of results.}
\revision{Our dataset contains} approximately 95K synchronized radar-camera-depth frames across 12 campus buildings, with over 40K frames under real smoke. 
\revision{We evaluate more than 25K frames from unseen buildings against six camera-only, radar-only, and radar-camera fusion baselines. \name{} obtains an MAE of 0.303\,m in clear scenes and 0.313\,m under smoke, with LPIPS of 0.126 and 0.137. Under heavy smoke, its Chamfer Distance is 0.104\,m$^2$, 40\% below GRT (0.173\,m$^2$) and 24\% below GRT+Image (0.137\,m$^2$); CaFNet and DA3 reach 3.503\,m$^2$ and 7.796\,m$^2$.}
Figure~\ref{fig-teaser} presents outputs across varying smoke densities against \revision{these} baselines.
Overall, we make the following contributions:
\begin{itemize}[leftmargin=*,itemsep=2pt,topsep=3pt]

    \item We introduce \name{}, the first system to apply diffusion-based generative refinement to single-frame (non-SAR) radar depth in egocentric 3D. \revision{\name{} produces} dense metric depth under visual degradation by grounding pretrained vision foundation model priors in radar geometry.

    \item We design a radar-to-depth translation stage that bridges the domain gap between 4D radar spectra and vision-domain representations, providing persistent metric grounding throughout the diffusion process and \revision{constraining refinement with the measured radar layout.}

    \item We propose a pixel-space residual visual guidance mechanism via a ControlNet-style adapter with zero-initialized skip connections. Trained with randomized degradation, it contributes spatial precision proportionally to available visual evidence without explicit mode switching.

    \item \revision{We validate building-disjoint generalization over 25K frames with real smoke and varying smoke densities}, achieving the best results across all metrics\revision{, with median MAE changing by 3.3\% from clear (0.303\,m) to smoke (0.313\,m)}. We will open-source the dataset, source code, and model weights.

\end{itemize}

\section{Related Work}
\label{sec:related}

Table~\ref{tab:comparison} summarizes the main differences from prior work.

\begin{table}[tb]
    \centering
    \scriptsize
    \setlength{\tabcolsep}{3.2pt}
    \renewcommand{\arraystretch}{0.95}
    \setlength{\aboverulesep}{0.25ex}
    \setlength{\belowrulesep}{0.25ex}

    \begin{tabular}{p{0.50\columnwidth}cccc}
        \toprule
        \textbf{Method} &
        \textbf{\shortstack{Robust\\Smoke}} &
        \textbf{\shortstack{3D\\Depth}} &
        \textbf{\shortstack{Single\\Frame}} &
        \textbf{\shortstack{Vision\\Prior}} \\
        \midrule

        \multicolumn{5}{l}{\textbf{Specialized Hardware \& Restoration}} \\
        Thermal / LiDAR~\cite{qwake_navigator,ouster_l2x_2021}
        & $\circ$ & \checkmark & $\circ$ & \ding{55} \\
        Inpainting~\cite{lugmayr2022repaint,saharia2022palette}
        & \ding{55} & \ding{55} & $\circ$ & \checkmark \\

        \midrule
        \multicolumn{5}{l}{\textbf{Multi-Modal Fusion}} \\
        Radar-Camera~\cite{sun2024cafnet,singh2023depth,wang2025tacodepth}
        & \ding{55} & \checkmark & \ding{55} & \ding{55} \\
        Radar-LiDAR/NIR~\cite{huang2025l4dr,palladin2024samfusion}
        & $\circ$ & \checkmark & \ding{55} & \ding{55} \\

        \midrule
        \multicolumn{5}{l}{\textbf{Radar-Only Perception}} \\
        SAR (PanoRadar)~\cite{panoradar}
        & \checkmark & \checkmark & \ding{55} & \ding{55} \\
        SAR (Millicam)~\cite{saadat2020millicam}
        & \checkmark & \checkmark & \ding{55} & \ding{55} \\
        Single-Frame (GRT)~\cite{grt}
        & \checkmark & \checkmark & \checkmark & \ding{55} \\
        Radar cGAN (MilliMap)~\cite{millimap}
        & \checkmark & \ding{55} & \ding{55} & \ding{55} \\

        \midrule
        \multicolumn{5}{l}{\textbf{Generative Models for Depth}} \\
        Monocular Diff.~\cite{marigold,lotus,yang2025da3d}
        & \ding{55} & \checkmark & $\circ$ & \checkmark \\
        Radar Diff. (RadarSFD)~\cite{zhao2025radarsfd}
        & \checkmark & \ding{55} & \checkmark & \checkmark \\

        \midrule
        \textbf{\name{} (Ours)}
        & \checkmark & \checkmark & \checkmark & \checkmark \\
        \bottomrule
    \end{tabular}
    \caption{\small{Related work summary.}}
    \vspace{-0.4in}
    \label{tab:comparison}
\end{table}

\noindent\textbf{Depth Under Visual Degradation.}
Thermal cameras~\cite{qwake_navigator} and multi-echo LiDAR~\cite{ouster_l2x_2021} extend optical-band sensing but remain attenuated by dense smoke~\cite{bijelic2020seeing} and provide no metric depth under uniform heating. Image restoration~\cite{cycle_gan_dehaze,guo2021_efficient_derain,liang2021swinir,zamir2021multi} and diffusion inpainting~\cite{lugmayr2022repaint,saharia2022palette} require residual visual signal; they fail when the optical path is blocked. These approaches remain coupled to optical availability, the failure regime \name{} targets.

\noindent\textbf{Multi-Modal Radar Fusion.}
Camera-radar~\cite{sun2024cafnet,wang2025tacodepth,long2021radar,singh2023depth} and radar-LiDAR/NIR stacks~\cite{huang2025l4dr,palladin2024samfusion,yang2025da3d} compensate for radar sparsity by treating the optical sensor as the primary feature source. This assumption breaks under degradation: CaFNet's Chamfer Distance rises from 0.174\,m$^2$ to 1.85\,m$^2$ under smoke in our evaluation, and LiDAR stacks are too bulky for compact mobile platforms. 
\revision{\name{} reverses this dependency: radar supplies the metric estimate, while RGB provides an auxiliary correction when usable.}

\noindent\textbf{Synthetic Aperture Radar.}
SAR achieves high resolution via coherent accumulation over sensor displacement~\cite{gao2021mimo,yanik2020development}. PanoRadar~\cite{panoradar} reaches LiDAR-comparable 3D quality through dense smoke via mechanical rotation; Millicam~\cite{saadat2020millicam} relaxes constraints with IMU trajectories. The fundamental limitation here is that SAR requires physical displacement ($\lambda/2 = 1.9$\,mm at 79\,GHz) and temporal accumulation, precluding per-frame depth on static platforms. 
\revision{\name{} instead uses learned priors to refine depth when controlled displacement or accurate relative pose is unavailable.}

\noindent\textbf{Single-Frame Radar Perception.}
Static and single-shot methods~\cite{zhou2025rise,prabhakara2023radarhd,millimap} and category-specific body reconstruction~\cite{xue2022m4esh,adhikari2022mishape} avoid motion requirements but do not generalize to free-form environments. 
Several RF and acoustic systems recover angle or depth through spatial encoding using metastructure augmentation ~\cite{garg2023sirius,garg2021owlet,bai2022spidr}.
GRT~\cite{grt} is the strongest general baseline, predicting 3D voxel occupancy from a single radar frame without SAR, but its resolution is bounded by the physical aperture (LPIPS\,$>$\,0.44 in our evaluation), and model scaling yields diminishing returns~\cite{grt}. \name{} addresses this \revision{aperture-imposed resolution limit} by grounding a pretrained diffusion backbone in GRT-style coarse depth.

\noindent\textbf{Generative Depth and Radar Perception.}
Marigold~\cite{marigold}, Lotus~\cite{lotus}, and DA3~\cite{da3} demonstrate that pretrained diffusion backbones encode rich geometric priors, but are entirely camera-dependent (DA3 MAE: 0.500\,m clear to 1.255\,m under smoke). In the radar domain, MilliMap~\cite{millimap} and diffusion-based BEV super-resolution~\cite{RAL24,ICRA24} operate in 2D. RadarSFD ~\cite{zhao2025radarsfd} adapts the Marigold paradigm to single-frame radar but remains in 2D BEV and does not address hallucination or degraded visual guidance. \name{} extends generative radar depth to egocentric 3D with a persistent metric anchor and a degradation-aware visual branch.

\begin{figure*}[t]
    \centering
    \vspace{-0.05in}
    \includegraphics[width=\textwidth]{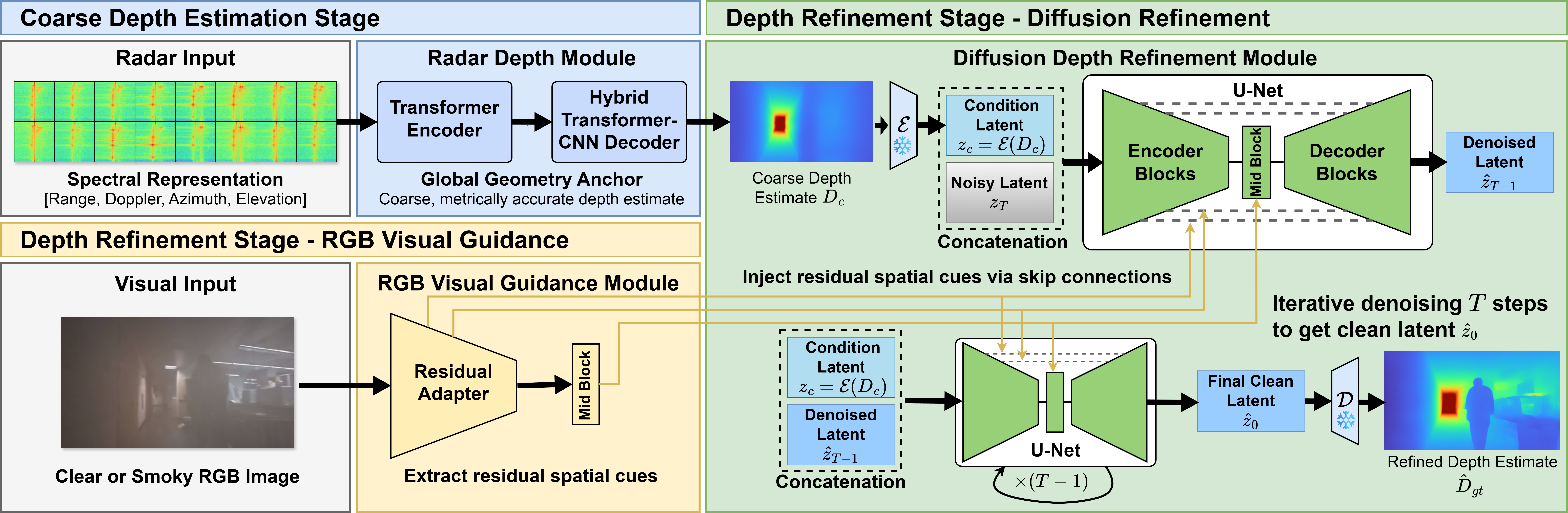}
    \caption{\small{System architecture of \name{}.
    The Radar Depth Module (Stage~1) maps the 4D radar spectrum to a coarse metric depth.
    The Diffusion Refinement Module (Stage~2) conditions a pretrained latent diffusion backbone on this radar depth, recovering fine structural detail using learned world priors.
    The RGB Visual Guidance Module (also Stage~2) injects pixel-space residual cues from the camera into the diffusion denoising process, providing spatial grounding when the image is reliable.}}
    \label{fig:pipeline}
\end{figure*}

\section{System Design}
\label{sec:design}

\subsection{Overview}
\label{sec:overview}

Our goal with \name{} is to generate dense, high-fidelity metric depth from a single radar frame in environments where cameras are partially or fully blinded by smoke or darkness. Achieving this requires more than just a capable radar model. While single-chip mmWave radars are robust to smoke, they are fundamentally limited in angular resolution due to small antenna arrays. Their raw measurements are coarse, sparse, and have ambiguities at object boundaries due to specular reflections. RGB cameras, on the other hand, are rich in structural detail but \revision{lose that detail rapidly under degraded visibility, and} they lack reliable metric scale even in clear conditions \cite{bijelic2020seeing,Guizilini2023ZeroDepth}. Therefore, neither modality alone delivers the capability that we need.

The key opportunity that we exploit in this paper is that large-scale diffusion models which are pretrained on web-scale image and depth data encode rich structural priors about how indoor scenes are organized, how surfaces terminate, how edges connect, how objects fit into space, etc. We believe these priors can supply the structural detail that radar physically cannot. The challenge, however, is that such models cannot operate directly on raw radar measurements - the domain gap from RF to RGB is too large. This motivates us to create our design pipeline. First, bridge the modality gap by translating radar into a coarse but metrically grounded depth estimate. Second, use pretrained generative priors to refine that estimate into a high-fidelity output. Finally, anchor the refinement to spatial reality using whatever residual visual cues the camera can still provide.

\begin{figure*}[t]
    \centering
    \setlength{\tabcolsep}{0pt}
    \renewcommand{\arraystretch}{1.0}
    \begin{tabular}{ccccc}
        \includegraphics[width=0.2\textwidth, valign=m]{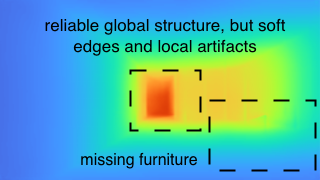} &
        \includegraphics[width=0.2\textwidth, valign=m]{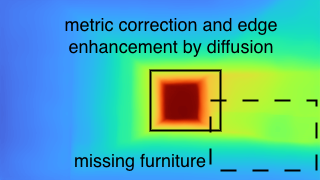} &
        \includegraphics[width=0.2\textwidth, valign=m]{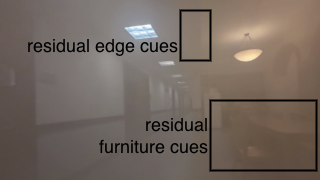} &
        \includegraphics[width=0.2\textwidth, valign=m]{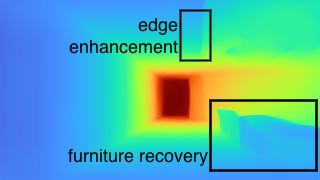} &
        \includegraphics[width=0.2\textwidth, valign=m]{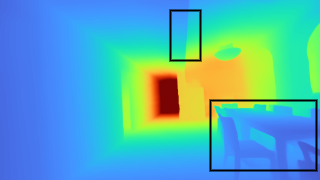} \\
        \multicolumn{1}{c}{{\fontsize{8}{9}\selectfont Output from Radar Model}} &
        \multicolumn{1}{c}{{\fontsize{8}{9}\selectfont Diffusion w/o Visual Guidance}} &
        \multicolumn{1}{c}{{\fontsize{8}{9}\selectfont Visual Cues}} &
        \multicolumn{1}{c}{{\fontsize{8}{9}\selectfont Diffusion w/ Visual Guidance}} &
        \multicolumn{1}{c}{{\fontsize{8}{9}\selectfont Ground Truth}} \\
    \end{tabular}
    \caption{\small{Intermediate outputs through the entire pipeline.}}
    \label{fig:system_flow}
\end{figure*}

\name{} is organized as a three-module pipeline running in two stages, as shown in Figure~\ref{fig:pipeline}. In Stage~1, the \textbf{Radar Depth Module} (\S\ref{sec:radar_module}) maps the 4D radar spectrum to a coarse metric depth image, establishing the geometric foundation for all subsequent processing. In Stage~2, the \textbf{Diffusion Depth Refinement Module} (\S\ref{sec:diffusion_module}) conditions a pretrained latent diffusion backbone on this radar depth, denoising from Gaussian noise into a structurally refined depth estimate. Running in parallel within Stage~2, the \textbf{RGB Visual Guidance Module} (\S\ref{sec:rgb_module}) injects pixel-space residual cues from the camera image into the diffusion process via a ControlNet-style adapter, providing spatial grounding when the camera is reliable and gracefully withdrawing when it is not. Throughout the pipeline, the radar depth controls global geometry; the diffusion model enriches that geometry using learned world priors; and the RGB branch corrects fine spatial detail placement only when trustworthy image evidence is available. Figure~\ref{fig:system_flow} traces an example prediction through all three stages, showing the cumulative contribution of each module.

\subsection{Radar Depth Module}
\label{sec:radar_module}

\noindent\textbf{Challenge.}
Recent single-frame radar depth methods, most notably GRT~\cite{grt}, have demonstrated significant progress by reformulating radar depth prediction as 3D voxel occupancy estimation from raw radar spectra, achieving for the first time foundational 3D scene understanding without SAR. \revision{Yet the achievable resolution remains bounded by the physical aperture.} Voxel occupancy projected to the image plane yields pixelated depth, and GRT itself shows that scaling model size yields diminishing returns~\cite{grt} - the bottleneck is not the model but the physical angular resolution of a single-chip radar. More importantly, the captured raw 4D radar spectra are entirely outside the training distribution of any pretrained vision foundation model, so the domain gap between radar and learned image priors must be bridged explicitly before generative refinement can be applied.

\noindent\textbf{Insight.}
Rather than treating the radar module as a complete solution, we design it as a modality translator. The goal is not to maximally recover depth from radar alone, but to produce a metrically grounded, spatially coherent depth image that lies within the distribution the downstream diffusion model can meaningfully condition on. This reframing changes the design target. Instead of predicting 3D voxel occupancy and projecting to depth, we learn a direct radar-to-depth-image mapping in image space, naturally producing dense predictions at image resolution and creating a conditioning signal the pretrained diffusion backbone can interpret. A further observation motivates the specific architecture choice: individual radar cells are inherently noisy and ambiguous, but adjacent cells reflecting from the same surface provide strong collective evidence for geometry. Cross-patch attention across the full radar field of view is therefore essential for extracting reliable geometric structure from sparse radar observations.

\noindent\textbf{Design.}
The radar depth module takes the 4D radar spectrum (range, Doppler, azimuth, elevation) as input and predicts a $128 \times 256$ coarse depth image via an encoder-decoder architecture.

\textit{Encoder.} The 4D radar cube is tokenized along the range and Doppler axes to extract azimuth-elevation patches, producing $N = 2048$ patch embeddings. Magnitude and phase channels are concatenated to preserve the complex physical relationships in the radar signal. A 4-layer transformer encoder applies self-attention across all patches, enabling global reasoning about surface coherence that no convolutional backbone can achieve at this input sparsity.

\textit{Decoder.} Encoded tokens are reshaped into a spatial feature grid and upsampled by a CNN decoder with bilinear interpolation and batch normalization. Transformers establish scene-level spatial awareness; convolutional upsampling exploits local structure bias for spatially coherent reconstruction.

\textit{Training objective.} We supervise with a combination of metric, perceptual, structural, and edge-aware losses:

\begin{equation}
\mathcal{L}_{\text{radar}} = \lambda_1 \| \hat{\mathbf{D}}_r - \mathbf{D}_{\text{gt}} \|_1 + \lambda_2 \, \text{LPIPS}(\hat{\mathbf{D}}_r, \mathbf{D}_{\text{gt}}) + \lambda_3 \, \mathcal{L}_{\text{SSIM}} + \lambda_4 \, \mathcal{L}_{\text{grad}}
\label{eq:radar_loss}
\end{equation}

where $\hat{\mathbf{D}}_r$ is the predicted coarse depth and $\mathbf{D}_{\text{gt}}$ is the ground truth. The $\ell_1$ term provides direct metric supervision, while LPIPS and SSIM encourage structural and perceptual consistency. The gradient loss $\mathcal{L}_{\text{grad}}$, motivated by its effectiveness in monocular depth estimation~\cite{da3}, explicitly penalizes depth discontinuity errors at surface boundaries,

\begin{equation}
\vspace{-0.05in}
\mathcal{L}_{\text{grad}} = \|\nabla_x \hat{\mathbf{D}}_r - \nabla_x \mathbf{D}_{\text{gt}}\|_1 + \|\nabla_y \hat{\mathbf{D}}_r - \nabla_y \mathbf{D}_{\text{gt}}\|_1
\label{eq:grad_loss}
\end{equation}

and proves critical for the downstream stage. Without $\mathcal{L}_{\text{grad}}$, the radar module produces smooth, blurred depth that lacks the structural sharpness needed to anchor diffusion refinement correctly. 
\revision{In our ablation study, we show that removing it raises MAE by 19--21\% and CD by 45--63\% across clear and smoke conditions (Table~\ref{tab:loss-ablations}).}

\begin{figure}[h]
    \vspace{-0.05in}
    \centering
    \setlength{\tabcolsep}{2pt}
    \renewcommand{\arraystretch}{1.0}
    \begin{tabular}{cc}
        \multicolumn{2}{c}{{\fontsize{8}{9}\selectfont \textbf{Simple scene}}} \\[2pt]
        \includegraphics[width=0.42\linewidth, valign=m]{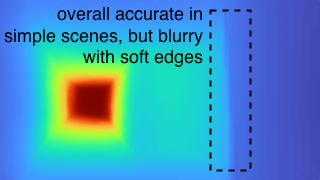} &
        \includegraphics[width=0.42\linewidth, valign=m]{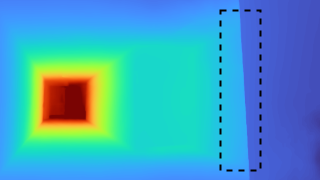} \\[10pt]
        \multicolumn{2}{c}{{\fontsize{8}{9}\selectfont \textbf{Complex scene}}} \\[2pt]
        \includegraphics[width=0.42\linewidth, valign=m]{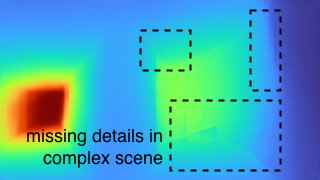} &
        \includegraphics[width=0.42\linewidth, valign=m]{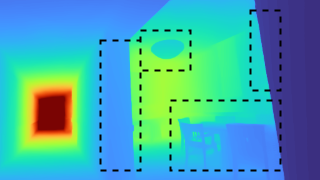} \\[4pt]
        {\fontsize{8}{9}\selectfont Output from Radar Module} &
        {\fontsize{8}{9}\selectfont Ground Truth} \\
    \end{tabular}
    \vspace{-0.05in}
    \caption{
    \small{Radar depth preserves global layout but lacks structural details, especially in complex scenes.}
    }
    \vspace{-0.1in}
    \label{fig:radar_output}
\end{figure}

\textit{Output characteristics.}
Figure~\ref{fig:radar_output} compares the coarse radar depth against ground truth across two scene types. In a simple scene (top row), the global layout is well recovered: walls, floors, and major objects sit at the correct distances. The bounding boxes highlight that only depth boundaries remain soft - a direct consequence of radar's physical angular resolution, not the model's capacity. This level of geometric fidelity is sufficient to anchor the next stage. In a complex scene (bottom row), however, the gap widens substantially. Clutter, small objects, and fine structural detail are missing or blurred; accurate global layout alone is not enough. This fidelity shortfall, which grows with scene complexity, is what makes the diffusion refinement module necessary.

\subsection{Diffusion Depth Refinement Module}
\label{sec:diffusion_module}

\noindent\textbf{Challenge.}
The radar aperture, rather than model capacity alone, causes the remaining structural gap. Edges, smooth surfaces, and small objects can fall below what a radar with few antennas resolves, and GRT reports diminishing returns from model scaling~\cite{grt}. Recovering this detail requires another source of geometric structure.

\noindent\textbf{Insight.}
Diffusion models pretrained on web-scale image and depth data learn a rich prior distribution over how indoor scenes are structured - how depth transitions at edges, how surfaces curve, how objects relate to one another spatially~\cite{ldm,marigold}. These priors encode exactly the structural knowledge that radar cannot physically recover. The key question is how to use them without introducing errors: a diffusion model applied without geometric constraints will generate depth that looks plausible but may be physically incorrect. Without grounding in the actual scene geometry, the model's prior can dominate, producing hallucinated structures at the wrong locations - a failure mode we examine in detail below. Our approach is to use diffusion not as a standalone estimator but as a conditional refinement operator, where the coarse radar depth persistently constrains every step of the denoising process, keeping the refinement within the radar-defined geometric manifold rather than allowing the prior to wander freely.

\noindent\textbf{Design.}
We use a latent diffusion model ~\cite{ldm} that performs denoising in a compressed latent space for computational efficiency.

\textit{Diffusion formulation.} Let $\mathbf{D}_{\text{gt}}$ denote the ground-truth depth, encoded as $\mathbf{z}_0 = \mathcal{E}(\mathbf{D}_{\text{gt}})$ by a pretrained frozen encoder $\mathcal{E}$. The forward process corrupts $\mathbf{z}_0$ over $T$ timesteps,
\begin{equation}
\vspace{-0.05in}
\mathbf{z}_t = \sqrt{\bar{\alpha}_t}\, \mathbf{z}_0 + \sqrt{1 - \bar{\alpha}_t}\, \boldsymbol{\epsilon}, \quad \boldsymbol{\epsilon} \sim \mathcal{N}(\mathbf{0}, \mathbf{I})
\vspace{-0.01in}
\end{equation}
where $\bar{\alpha}_t$ follows a predefined noise schedule, and the model learns to reverse this process by predicting $\boldsymbol{\epsilon}$ at each step.

\begin{figure}[h]
    \vspace{-0.05in}
    \centering
    \setlength{\tabcolsep}{0pt}
    \renewcommand{\arraystretch}{1.0}
    \begin{tabular}{c}
        \includegraphics[width=0.99\linewidth, valign=m]{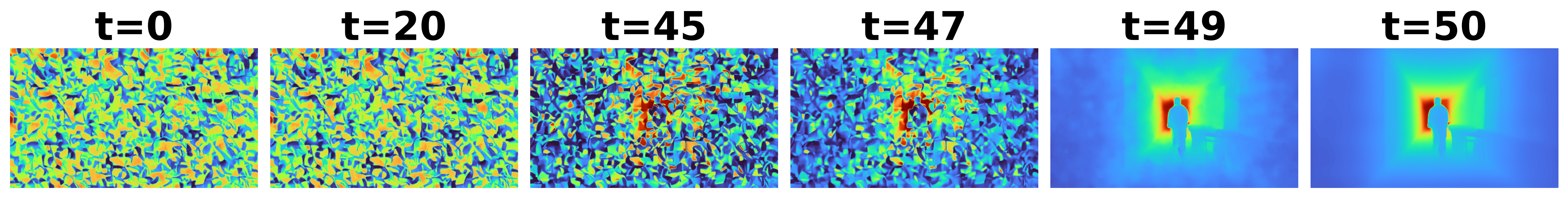} \\[6pt]
        \includegraphics[width=0.99\linewidth, valign=m]{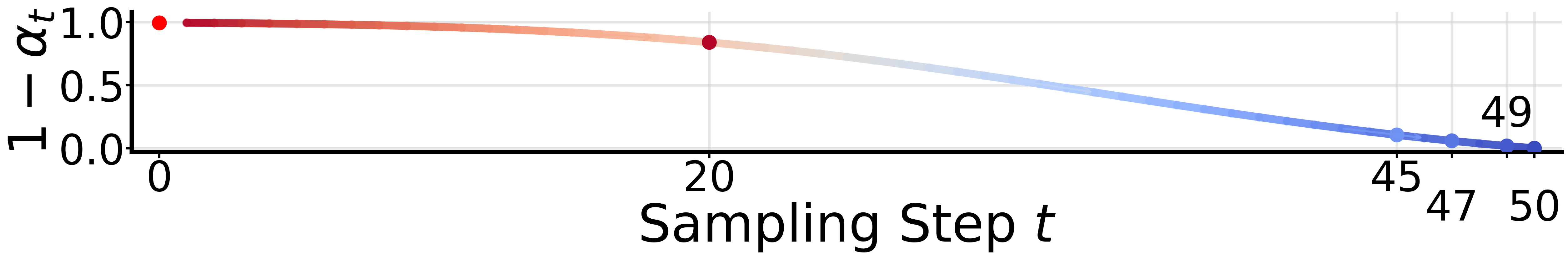}
    \end{tabular}
    \vspace{-0.05in}
    \caption{
    Radar conditioning iteratively denoises scene layout, while later steps recover finer structural detail.
    }
    \vspace{-0.05in}
    \label{fig:system_ddim_sampling}
\end{figure}

\textit{Radar-conditioned denoising.} The coarse radar depth $\hat{\mathbf{D}}_r$ is encoded as a conditioning latent $\mathbf{z}_c = \mathcal{E}(\hat{\mathbf{D}}_r)$, then concatenated channel-wise with the noised latent $\mathbf{z}_t$ to form an 8-channel input to the denoising U-Net $\epsilon_\theta$,
\begin{equation}
\hat{\boldsymbol{\epsilon}} = \epsilon_\theta\!\left([\mathbf{z}_c \,;\, \mathbf{z}_t],\, t\right)
\label{eq:denoising}
\end{equation}
This makes the radar geometry visible to every layer of the network at every denoising step. Rather than a one-time initialization, $\mathbf{z}_c$ acts as a persistent geometric anchor throughout the denoising trajectory - the model cannot ``forget'' the radar layout at any point during refinement. Early denoising steps recover coarse scene structure consistent with the radar conditioning; later steps progressively sharpen fine geometric detail, all within the radar-defined metric frame. Figure~\ref{fig:system_ddim_sampling} illustrates this progression: within the first few steps, the radar conditioning pulls the sample toward the correct scene layout; subsequent steps refine structure and detail without departing from the established geometry.

Radar conditioning also provides two compounding benefits worth noting. First, conditioning the diffusion model on a depth image rather than raw radar spectra substantially reduces the domain gap the pretrained model faces. Second, because $\mathbf{z}_c$ is present at every denoising step, the final output remains dominated by radar geometry and inherits radar's robustness to visual degradation - a property that carries all the way to inference.

\begin{figure}[h]
    \centering
    \vspace{-0.05in}
    \setlength{\tabcolsep}{2pt}
    \renewcommand{\arraystretch}{1.0}
    \begin{tabular}{ccc}
        \multicolumn{3}{c}{{\fontsize{8}{9}\selectfont \textbf{Refinement: diffusion adds missing detail correctly}}} \\
        \includegraphics[width=0.3\linewidth, valign=m]{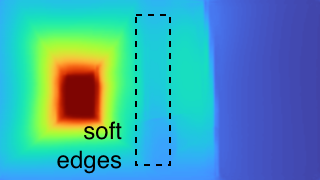} &
        \includegraphics[width=0.3\linewidth, valign=m]{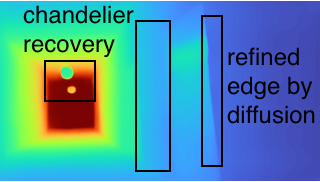} &
        \includegraphics[width=0.3\linewidth, valign=m]{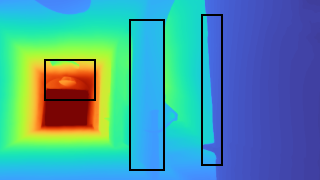} \\[8pt]
        \multicolumn{3}{c}{{\fontsize{8}{9}\selectfont \textbf{Hallucination: plausible structure at the wrong location}}} \\
        \includegraphics[width=0.3\linewidth, valign=m]{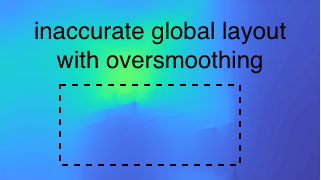} &
        \includegraphics[width=0.3\linewidth, valign=m]{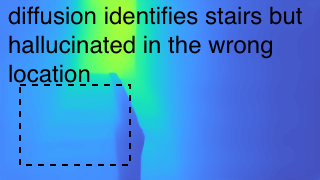} &
        \includegraphics[width=0.3\linewidth, valign=m]{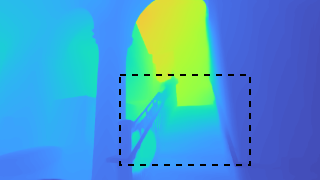} \\[4pt]
        {\fontsize{8}{9}\selectfont Radar Module} &
        {\fontsize{8}{9}\selectfont Diffusion w/o Visual} &
        {\fontsize{8}{9}\selectfont Ground Truth} \\
    \end{tabular}
    \vspace{-0.05in}
    \caption{
    Diffusion recovers missing detail when radar geometry is reliable, but hallucinates structures under radar geometric ambiguity.
    }
    \vspace{-0.05in}
    \label{fig:diffusion_hallucination}
\end{figure}

\textit{When priors help and when they hallucinate.}
Figure~\ref{fig:diffusion_hallucination} illustrates both regimes side by side. In the refinement case (top row), diffusion adds detail that radar physically cannot recover: a chandelier and fine ceiling structure are accurately inserted, closely matching ground truth. In the hallucination case (bottom row), the radar output has an over-smoothed global layout that leaves the scene geometry underspecified. The diffusion model, drawing on its prior, correctly infers that a staircase should exist - this inference is geometrically plausible - but places it on the wrong side of the scene compared to ground truth. The prior knows what structures belong indoors; it does not know where they belong in this specific scene. 
\revision{Average pixel metrics such as $\ell_1$ and SSIM can underweight this spatial misplacement, although it directly affects downstream use.}
This limitation directly motivates the RGB visual guidance module.

\textit{Training objective.} We train the diffusion module with a joint latent-space and pixel-space loss,

\begin{equation}
\begin{aligned}
\mathcal{L}_{\text{diff}} =
&\underbrace{\|\hat{\boldsymbol{\epsilon}} - \boldsymbol{\epsilon}\|_2^2}_{\text{denoising}}
+ \lambda_5 \|\hat{\mathbf{D}} - \mathbf{D}_{\text{gt}}\|_1 \\
&+ \lambda_6\, \text{LPIPS}(\hat{\mathbf{D}}, \mathbf{D}_{\text{gt}})
+ \lambda_7 \mathcal{L}_{\text{SSIM}}
+ \lambda_8 \mathcal{L}_{\text{grad}}
+ \lambda_9 \mathcal{L}_{\text{3D}}
\end{aligned}
\label{eq:diff_loss}
\end{equation}
where $\hat{\mathbf{D}} = \mathcal{D}(\hat{\mathbf{z}}_0)$ is the depth decoded from the estimated clean latent $\hat{\mathbf{z}}_0$. The denoising term trains the conditional process in latent space; the pixel-space terms ensure a metrically correct, edge-sharp output.

\textit{3D reconstruction loss.} Pixel-level and perceptual losses do not fully capture geometric correctness in 3D. A depth map that appears locally plausible in 2D can exhibit substantial misalignment when back-projected to 3D camera coordinates. We address this with an explicit 3D geometric loss. Each predicted depth value is unprojected into 3D camera coordinates using the pinhole camera model, and the loss penalizes spatial misalignment across all valid pixels,
\begin{equation}
\vspace{-0.02in}
\mathcal{L}_{\text{3D}} = \frac{1}{N}\sum_{i=1}^{N} \left( |x_i^{\text{pred}} - x_i^{\text{gt}}| + |y_i^{\text{pred}} - y_i^{\text{gt}}| + |z_i^{\text{pred}} - z_i^{\text{gt}}| \right)
\label{eq:3d_loss}
\vspace{-0.02in}
\end{equation}
This term encourages the refinement module to preserve the radar-defined metric structure in 3D, not just optimize 2D appearance.

\subsection{RGB Visual Guidance Module}
\label{sec:rgb_module}

\noindent\textbf{Challenge.}
The hallucination failure shown in Figure~\ref{fig:diffusion_hallucination} is a spatial ambiguity problem. The diffusion prior knows that indoor structures like staircases exist; without a spatial reference it cannot reliably determine where in this particular scene they belong. A camera image - even a heavily smoke-degraded one - encodes exactly this spatial reference from the same viewpoint at full pixel resolution. Yet the camera is also the first modality to fail as visibility degrades. \revision{A naive fusion strategy that treats RGB as a symmetric input would inherit camera failure modes in exactly the settings where radar depth estimation is most needed.}

\noindent\textbf{Insight.}
Our central observation is that even a degraded image can have useful features. Under heavy smoke, partial edges, surface boundaries, and structural transitions often survive as sparse but informative residual cues. The challenge is extracting these cues without losing them. Feeding the degraded image through a pretrained latent encoder $\mathcal{E}$ is problematic because $\mathcal{E}$ was trained on clean images and has no mechanism to distinguish residual signal from smoke-induced noise; it tends to suppress exactly the surviving sparse structure during compression. Operating in pixel space avoids this bottleneck, preserving whatever spatial evidence remains at full resolution.

Interestingly, a fully occluded image like a black or white image with no structural information is not a failure. It is a signal for camera's confidence score. \revision{When the visual branch receives a completely occluded input, its contribution to the denoising and sampling process should become negligible, gracefully deferring to the radar-conditioned diffusion path.} Rather than encoding this as an explicit rule, we design the architecture and training procedure so that this behavior emerges from our diffusion backbone. The visual guidance branch thus behaves as an asymmetric contributor- it adds spatial precision under favorable conditions and withdraws gracefully as the conditions worsen due to smoke.

\noindent\textbf{Design.}
We adopt a ControlNet-style architecture~\cite{controlnet} operating in pixel space rather than latent space.

\textit{Pixel-space residual encoder.} A trainable copy of the U-Net encoder blocks processes the RGB image $\mathbf{I} \in \mathbb{R}^{3 \times H \times W}$ at full resolution without latent compression. The encoder extracts multi-scale feature maps $\{\mathbf{f}_l\}_{l=1}^{L}$ that capture hierarchical spatial structure at progressively finer scales. These features are injected into the corresponding decoder blocks and mid-block of the main denoising U-Net via zero-initialized additive skip connections $\mathbf{h}_l' = \mathbf{h}_l + \mathbf{f}_l$ where $\mathbf{h}_l$ is the main U-Net decoder activation at layer $l$. 
\revision{Zero initialization makes the visual branch contribute nothing at the start of training, so the radar-conditioned backbone retains its behavior while the adapter learns. The branch then adjusts local detail placement while global geometry stays anchored by the radar conditioning latent $\mathbf{z}_c$, which is present at every denoising step.}

\begin{figure}[h]
    \vspace{-0.05in}
    \centering
    \setlength{\tabcolsep}{0pt}
    \renewcommand{\arraystretch}{1.0}
    \begin{tabular}{cccc}
        \multicolumn{4}{c}{{\fontsize{8}{9}\selectfont \textbf{Clear Visibility}}} \\
        \includegraphics[width=0.23\linewidth, valign=m]{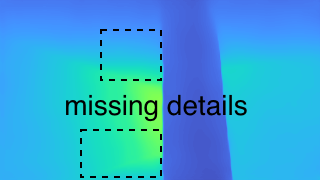} &
        \includegraphics[width=0.23\linewidth, valign=m]{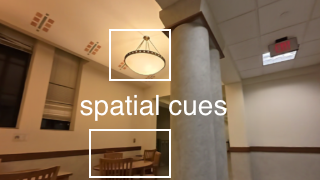} &
        \includegraphics[width=0.23\linewidth, valign=m]{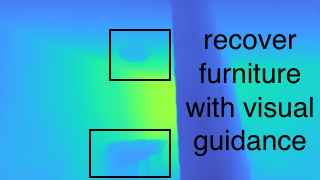} &
        \includegraphics[width=0.23\linewidth, valign=m]{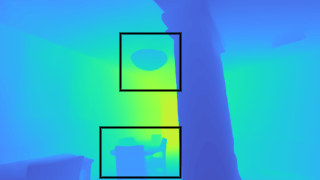} \\[6pt]
        \multicolumn{4}{c}{{\fontsize{8}{9}\selectfont \textbf{Heavy Smoke}}} \\
        \includegraphics[width=0.23\linewidth, valign=m]{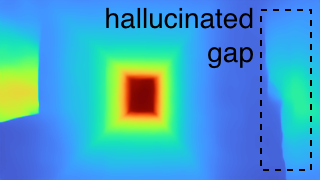} &
        \includegraphics[width=0.23\linewidth, valign=m]{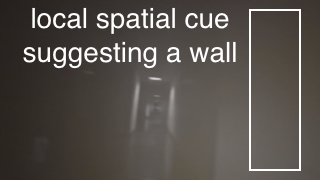} &
        \includegraphics[width=0.23\linewidth, valign=m]{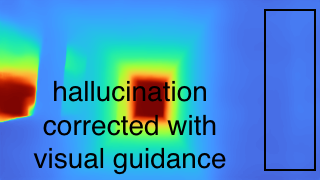} &
        \includegraphics[width=0.23\linewidth, valign=m]{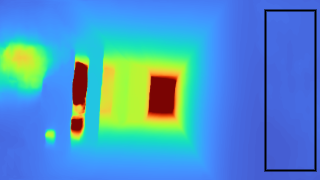} \\[6pt]
        \multicolumn{4}{c}{{\fontsize{8}{9}\selectfont \textbf{Zero Visibility}}} \\
        \includegraphics[width=0.23\linewidth, valign=m]{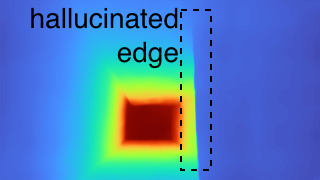} &
        \includegraphics[width=0.23\linewidth, valign=m]{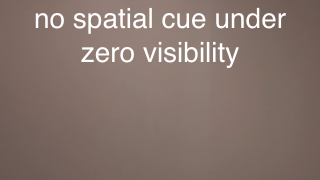} &
        \includegraphics[width=0.23\linewidth, valign=m]{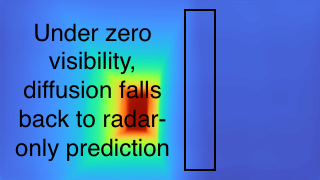} &
        \includegraphics[width=0.23\linewidth, valign=m]{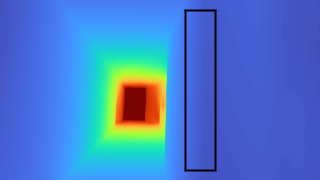} \\[4pt]
        {\fontsize{8}{9}\selectfont Diffusion w/o Visual} &
        {\fontsize{8}{9}\selectfont Visual Input} &
        {\fontsize{8}{9}\selectfont Diffusion w/ Visual} &
        {\fontsize{8}{9}\selectfont Ground Truth} \\
    \end{tabular}
    \vspace{-0.05in}
        \caption{RGB guidance improves detail when useful visual features remain \revision{and falls back to radar-only in zero-visibility.}}
    \vspace{-0.05in}
    \label{fig-system-full}
\end{figure}

\textit{Graceful degradation across visibility regimes.}
Figure~\ref{fig-system-full} shows the module's behavior across three representative conditions. Under clear visibility (top row), the camera provides dense, high-contrast features. The visual guidance branch uses these to \revision{resolve} fine spatial detail - furniture locations, chair boundaries, object placements - that the diffusion model alone cannot reliably localize, as highlighted by the bounding box annotations. Under heavy smoke (middle row), the image is severely degraded but not blank. Residual edges and partial surface transitions survive, and these sparse cues are sufficient for the branch to correct the spatial hallucination from the diffusion-only path, directing the staircase to its correct location without introducing new artifacts. Under zero visibility (bottom row), the camera input is fully occluded and carries no spatial signal. 
\revision{Under zero visibility, the skip connections become small, and the full output closely follows the radar-conditioned diffusion prediction. \S\ref{sec:graceful} quantifies the corresponding output-level convergence across measured smoke density.}

\textit{Randomized smoke augmentation.}
\revision{Training exposes the visual adapter to clean images, synthetic fog, real smoke, and full occlusion. Synthetic fog follows Beer--Lambert attenuation $T=\exp(-\beta d)$ with randomized $\beta$ and Perlin-modulated spatial density; real-smoke frames retain the physical image degradation measured during collection; and full occlusion is simulated with uniform overexposed images. These inputs train the adapter across the evaluated visibility regimes without imposing an explicit reliability gate. Section~\ref{sec:implementation} reports the exact sampling distribution.}

\section{Implementation}
\label{sec:implementation}

\subsection{Dataset}
\label{sec:dataset}

We train on two sources. IQ-1M~\cite{grt} provides synchronized radar, camera, and LiDAR measurements across indoor, outdoor, and bike scenes; we use only the indoor subset to pretrain GRT and our radar depth module.

\begin{figure}[h]
    \vspace{-0.05in}
    \centering
    \setlength{\tabcolsep}{2pt}
    \renewcommand{\arraystretch}{1.0}
    \begin{tabular}{cc}
        \includegraphics[width=0.45\linewidth, valign=m]{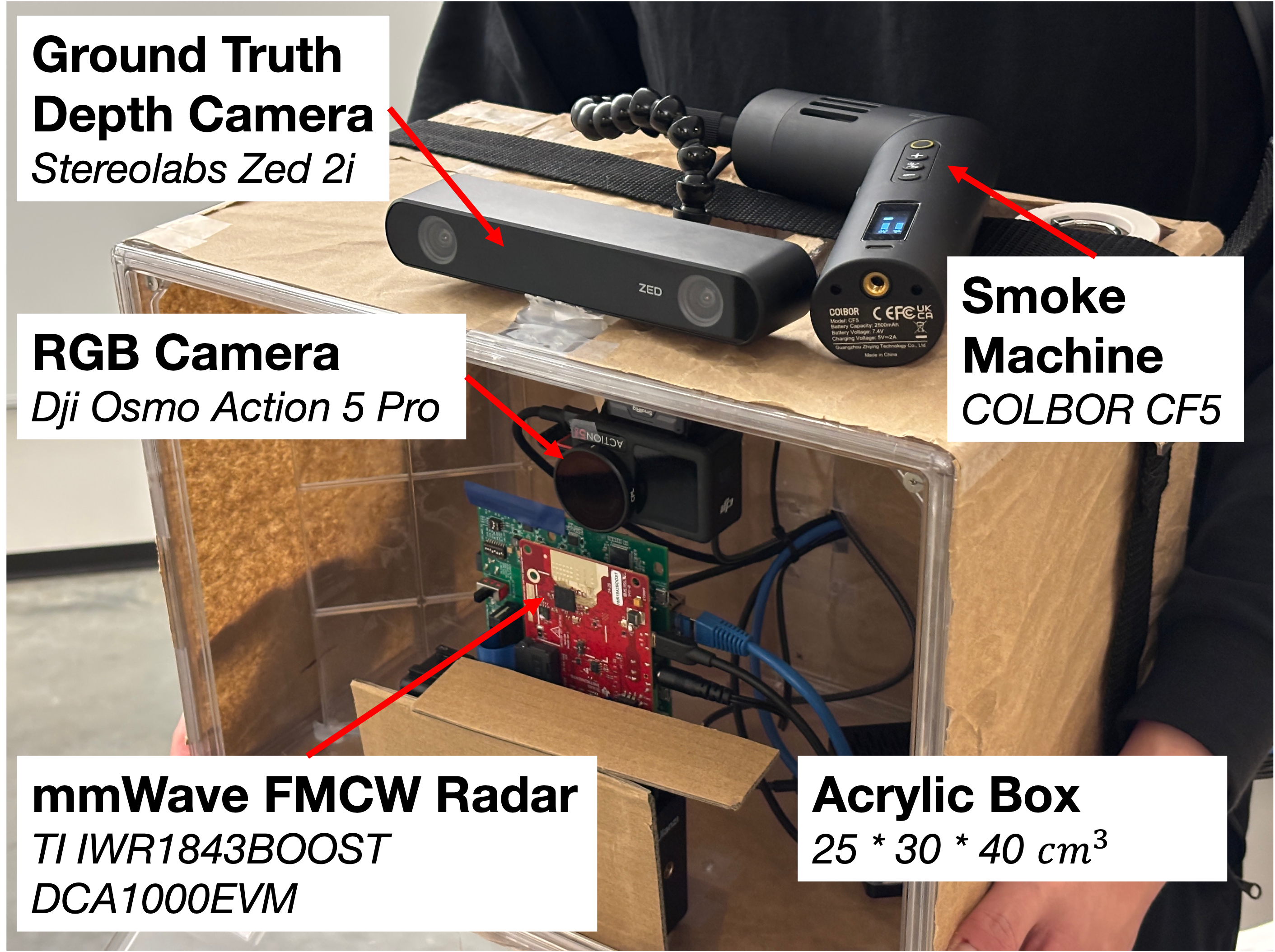} &
        \includegraphics[width=0.45\linewidth, valign=m]{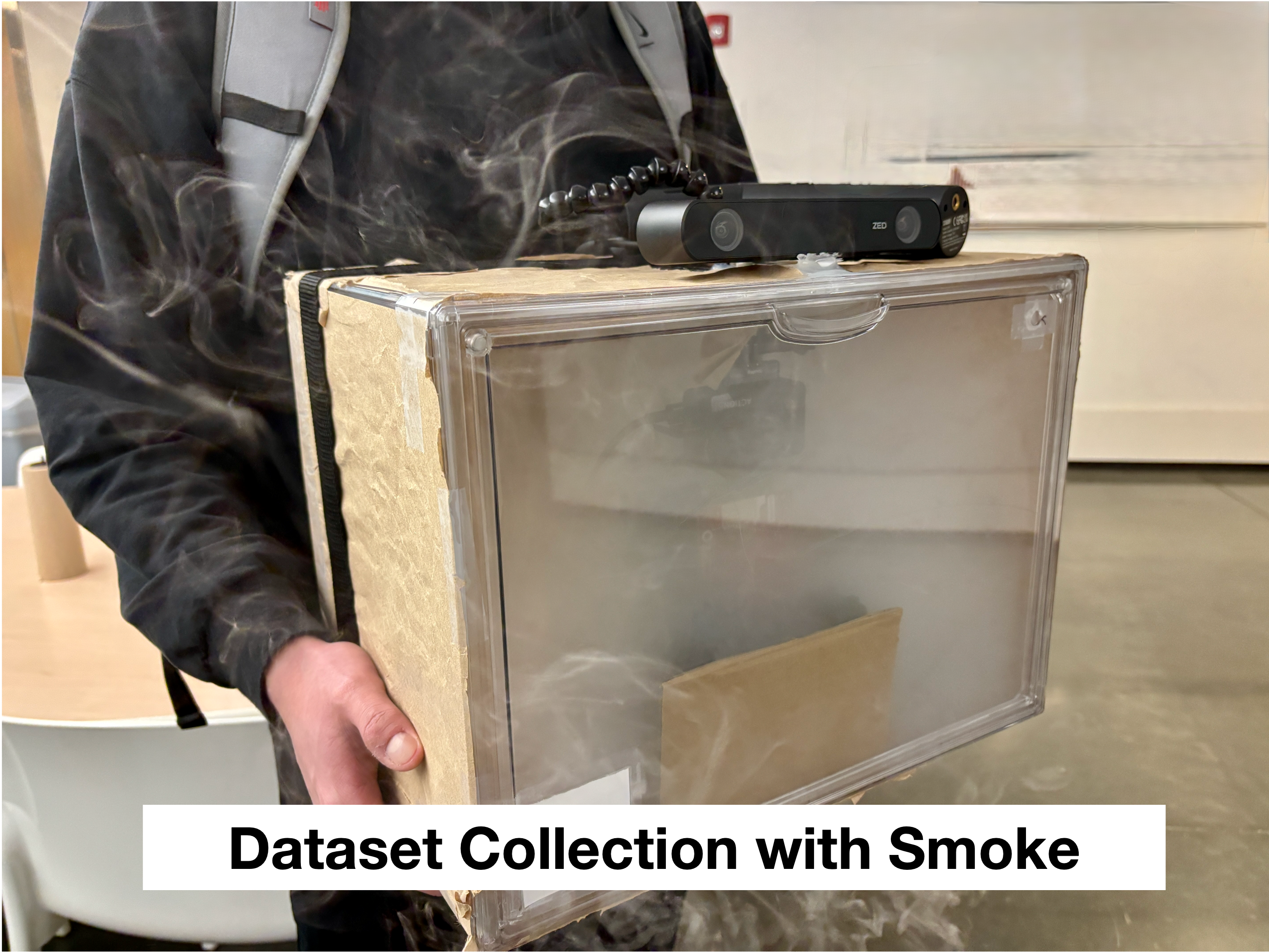} \\
    \end{tabular}
    \vspace{-0.05in}
    \caption{\small{Our prototype and data collection setup in smoke.}}
    \vspace{-0.05in}
    \label{fig:prototype}
\end{figure}

Our self-collected dataset is captured using the prototype shown in Figure~\ref{fig:prototype}: a 77\,GHz mmWave FMCW radar (TI IWR1843BOOST + DCA1000EVM)~\cite{1843boost, dca1000evm} for raw I/Q acquisition, a DJI Action~5~Pro~\cite{action5pro} for smoke-degraded RGB, and a Stereolabs ZED~2i stereo camera~\cite{zed2i} for reference RGB and ground-truth depth. The radar operates at 10\,Hz; both cameras record 720p at 30\,FPS. A MAX30105 IR particle sensor~\cite{max30105} measures smoke density during evaluation sequences. All modalities are synchronized offline. The dataset contains approximately 95K synchronized frames, including more than 40K frames are captured under real smoke at varying densities, across 12 campus buildings. For evaluation, we adopt a building-level split. All test sequences come from buildings that are entirely disjoint from the training set. It helps us evaluate a realistic measure of generalization to unseen indoor environments.

\revision{\noindent\textit{Synchronization.} All frames are timestamped at acquisition and aligned against the lowest-rate sensor, the radar at 10\,Hz. Each radar frame is matched to its nearest ZED frame, then the nearest DJI frame to that ZED timestamp. Both steps are one-to-one within 50\,ms, and frames failing either synchronization are dropped.}

\revision{\noindent\textit{Calibration.} Using 800 synchronized checkerboard pairs, we calibrate the DJI fisheye model and estimate a fixed homography to the rectified ZED left-camera view. DJI frames are undistorted, warped, cropped to the shared field of view, and resized. Radar and ZED are rigidly co-mounted; their fixed cross-modal alignment is learned from paired data in this reference view.}

\revision{\noindent\textit{Ground truth under smoke.} Smoke is confined to a transparent acrylic enclosure containing the DJI camera, while the stereo ZED observes the same scene from outside along a clear optical path. Reference depth is therefore independent of smoke density.}

\subsection{Radar Preprocessing}
\label{sec:radar_preproc}

For each radar frame, raw I/Q measurements are first reorganized into a virtual MIMO array. We then apply FFT-based processing along the fast-time, slow-time, azimuth, and elevation dimensions to produce a 4D complex-valued spectrum spanning range, Doppler, azimuth, and elevation. The complex spectrum is converted to a real-valued tensor by separating magnitude and phase into two channels; the magnitude channel is scaled by $10^{-3}$ and the phase channel is normalized to $[0,1]$ for numerical stability during training.

\subsection{Training Setup}
\label{sec:training_config}
\name{} is trained in three sequential stages so that each module learns its role without interference.
All stages use Adam~\cite{kingma2014adam} with an initial learning rate of $1\times10^{-4}$, weight decay $1\times10^{-2}$, and FP16 mixed-precision training.

\textit{Stage~1.} The radar depth module is trained independently as a radar-to-depth predictor.

\textit{Stage~2.} The radar depth module is frozen. The autoencoder ($\mathcal{E}$, $\mathcal{D}$) uses TAESD weights and is also frozen; the denoising U-Net is initialized from Marigold~\cite{marigold}. We use the standard DDPM noise schedule with $T = 1{,}000$ timesteps~\cite{ddpm}.

\revision{\textit{Stage~3.} We freeze the radar and diffusion modules and train only the ControlNet encoder and zero-initialized skip connections. Real-smoke frames are used without additional degradation. For each clear-condition training sample, the RGB input is unmodified with probability 60\%, synthetically fogged with probability 30\%, or fully occluded with probability 10\%. Synthetic fog uses $T=\exp(-\beta d)$ with $\beta\in[0.08,0.32]$\,m$^{-1}$ and Perlin-modulated density; full occlusion uses a uniform intensity sampled from 200--255. These probabilities remain fixed throughout training, with no epoch-level dropout schedule, and no reliability estimator gates the branch.}

\textit{Inference Noise Scheduler.} At inference, we replace DDPM with DDIM sampling~\cite{ddim} with 8 steps, initialized from the same noise schedule used during training.
\vspace{-0.1in}

\section{Evaluation}
\label{sec:eval}

\subsection{Evaluation Setup}
\label{sec:setup}
\revision{We evaluate \name{} in clear and smoke-filled indoor scenes using a building-disjoint split: no test building appears in training. The test set contains more than 25{,}000 frames. We use the infrared reading from a MAX30105 sensor~\cite{max30105} as a proxy for smoke density and group frames as light ($\mathrm{IR}<2000$), medium ($2000\leq\mathrm{IR}<4000$), or heavy ($\mathrm{IR}\geq4000$).}
Code and datasets are available at \textcolor{blue}{\url{https://phi-lab-rice.github.io/GRADE}}.

\begin{figure*}[t]
    \vspace{-0.05in}
    \centering
    \setlength{\tabcolsep}{1pt}
    \renewcommand{\arraystretch}{0.9}
    \includegraphics[width=0.5\linewidth]{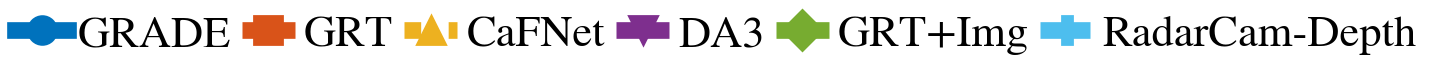}\\
    \begin{tabular}{@{}ccccc@{}}
        \includegraphics[width=0.248\textwidth, valign=m]{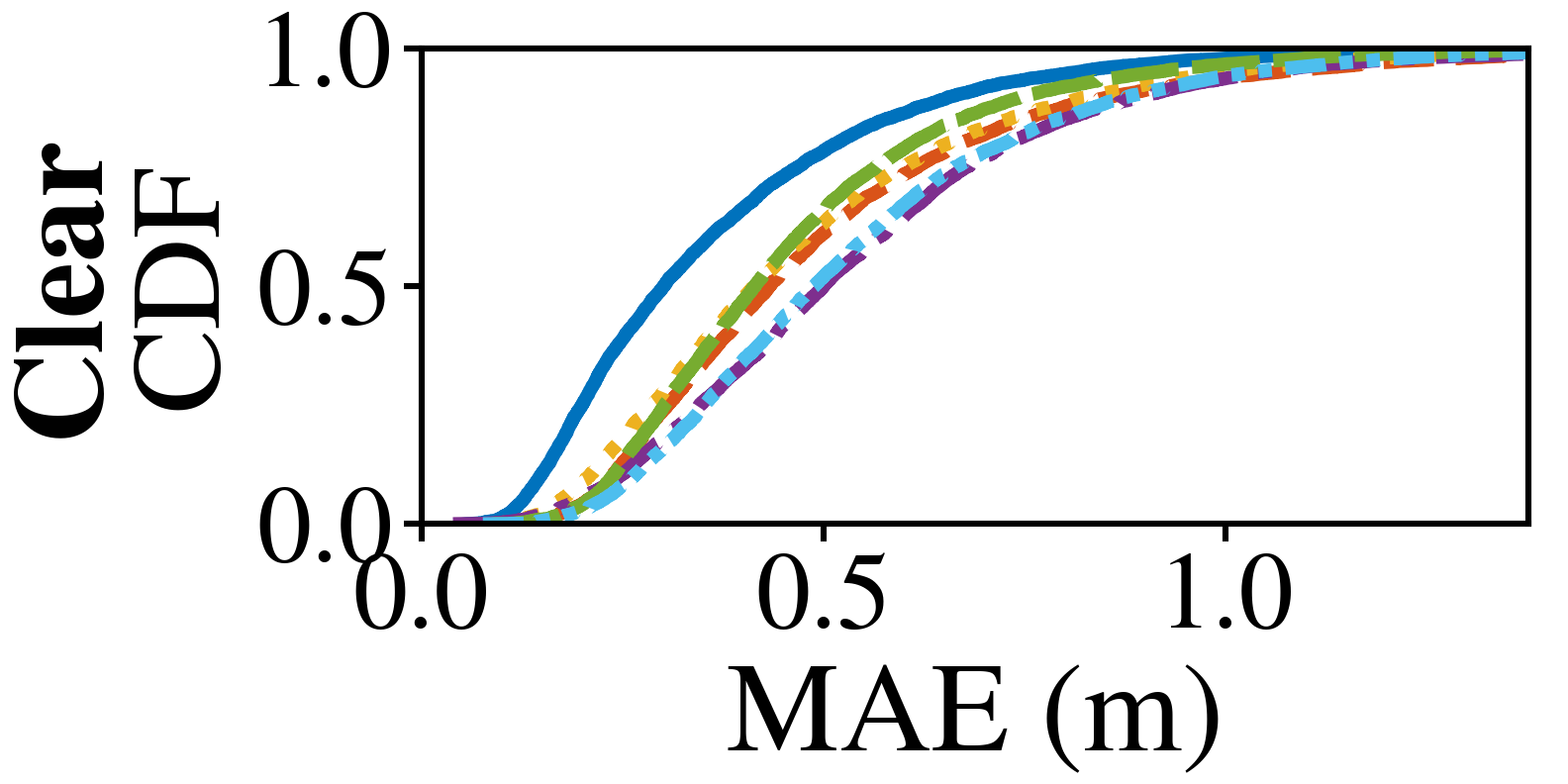} &
        \includegraphics[width=0.248\textwidth, valign=m]{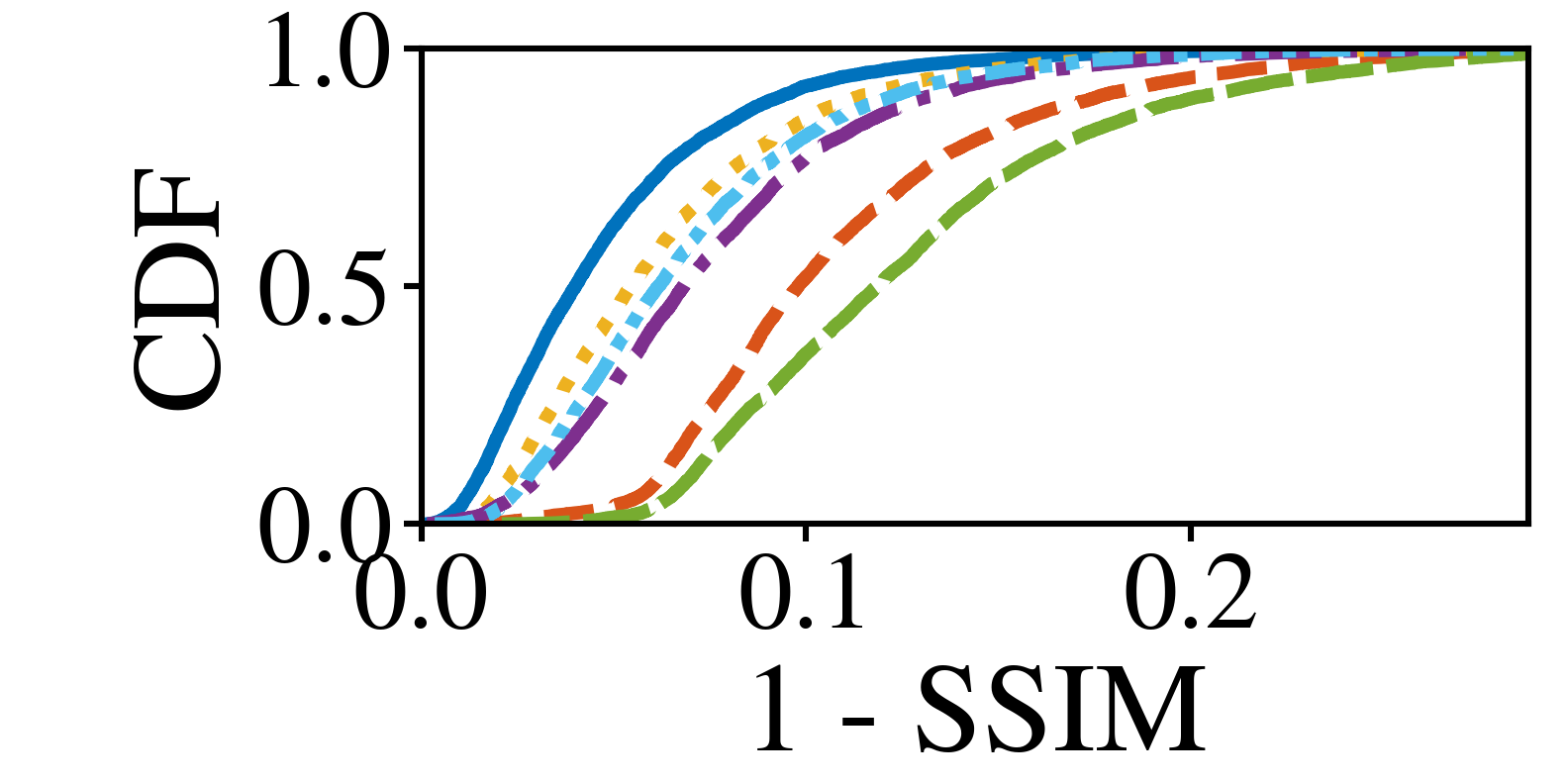} &
        \includegraphics[width=0.248\textwidth, valign=m]{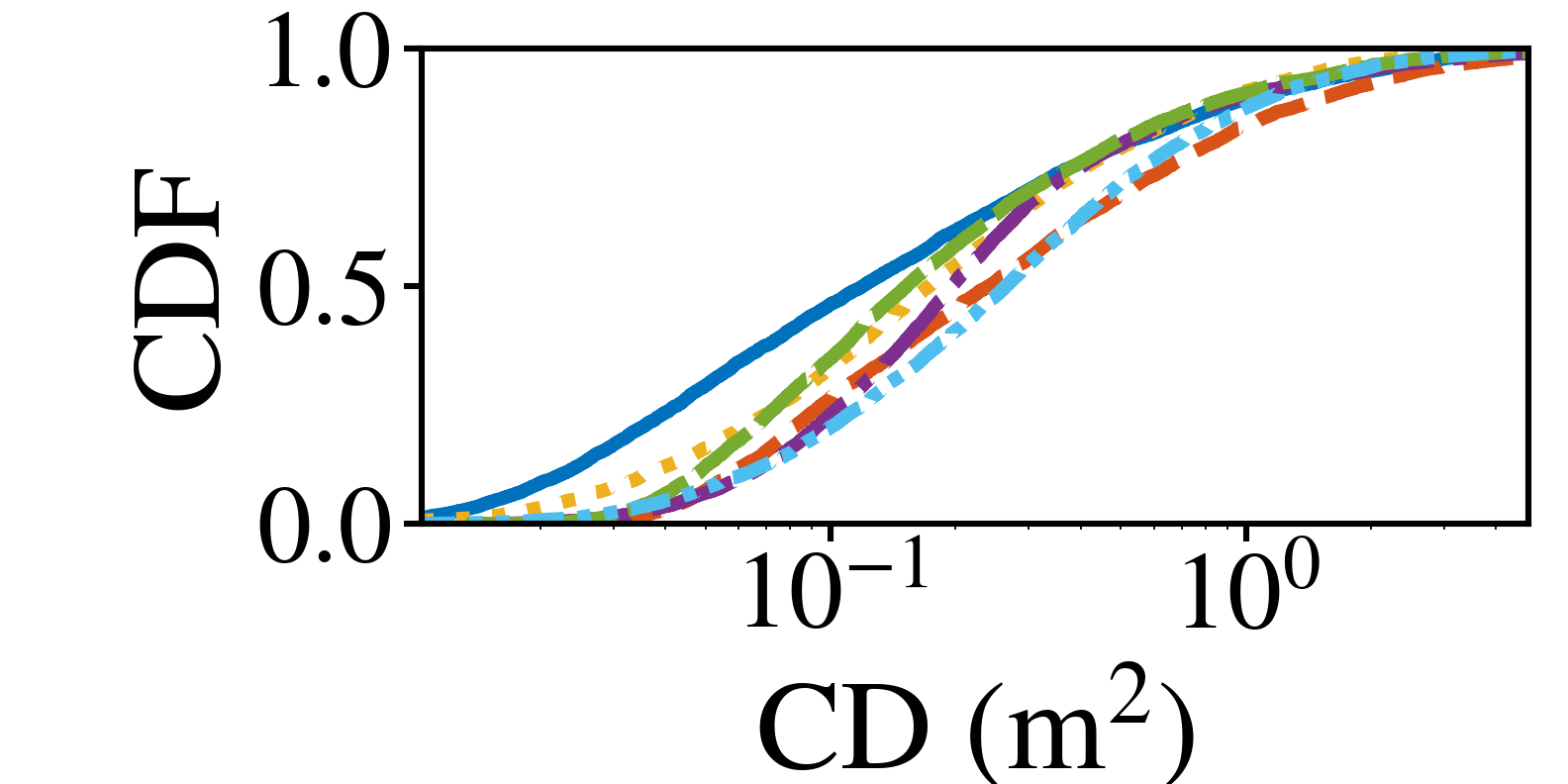} &
        \includegraphics[width=0.248\textwidth, valign=m]{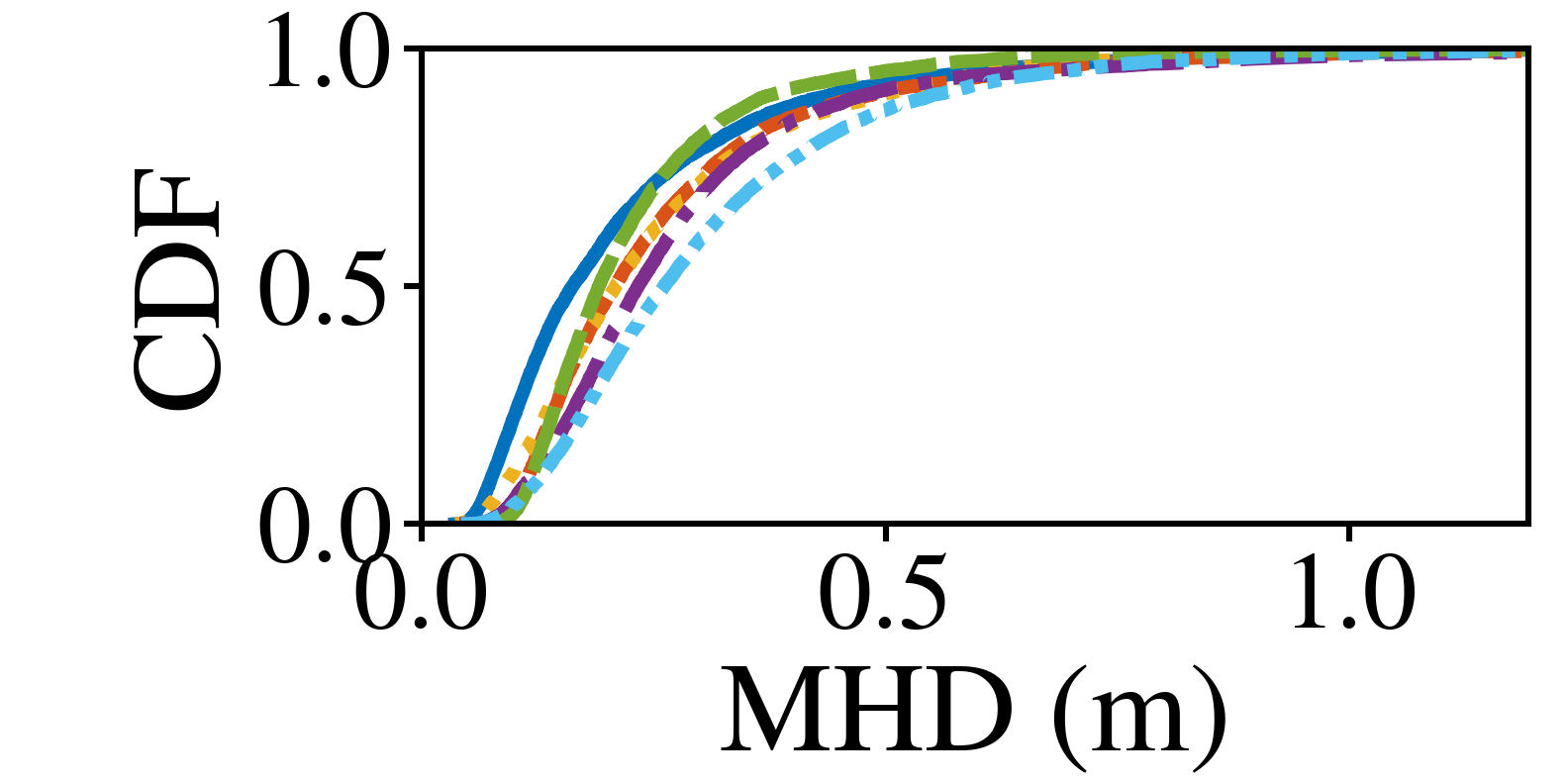} \\

        \includegraphics[width=0.248\textwidth, valign=m]{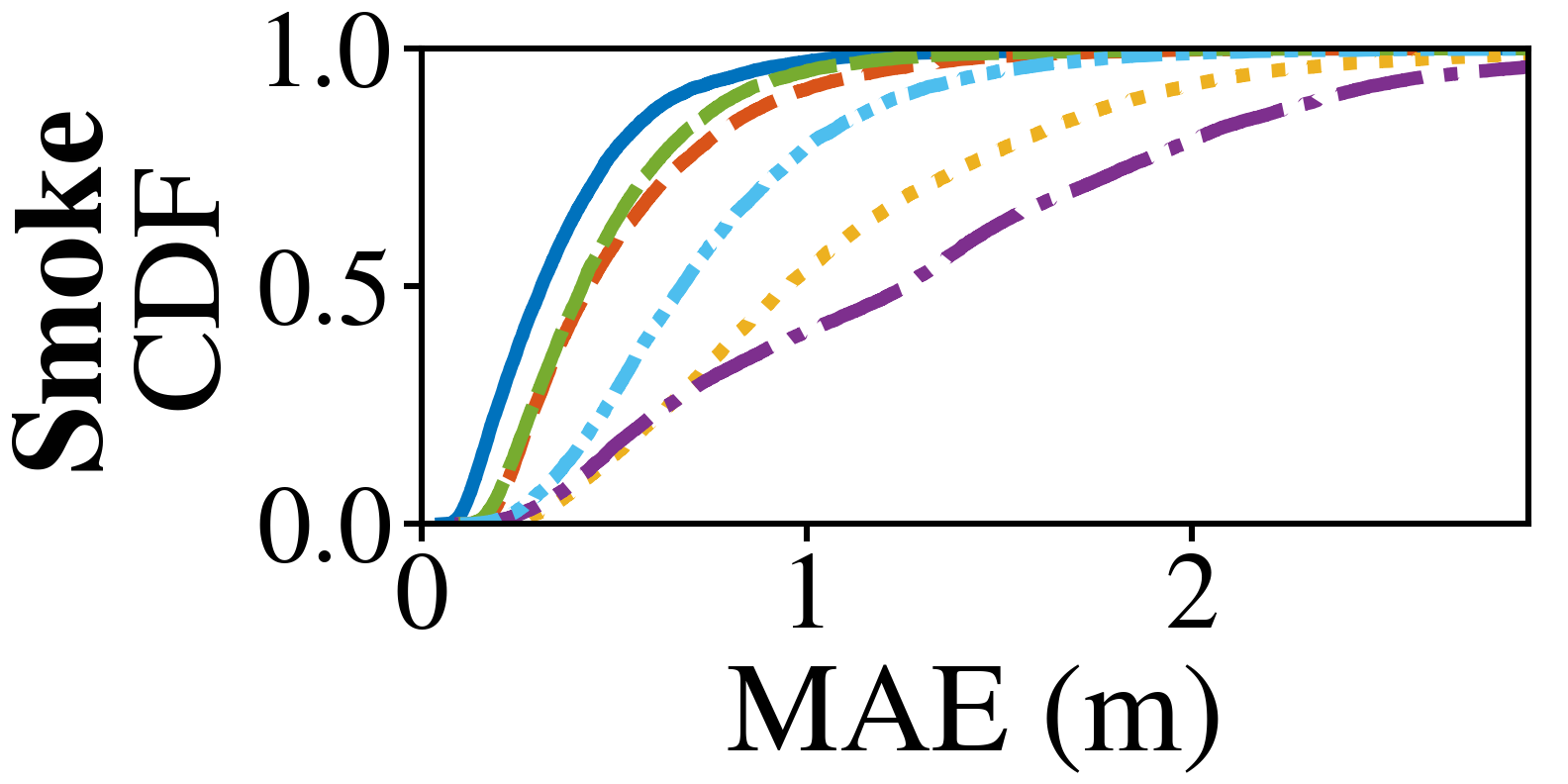} &
        \includegraphics[width=0.248\textwidth, valign=m]{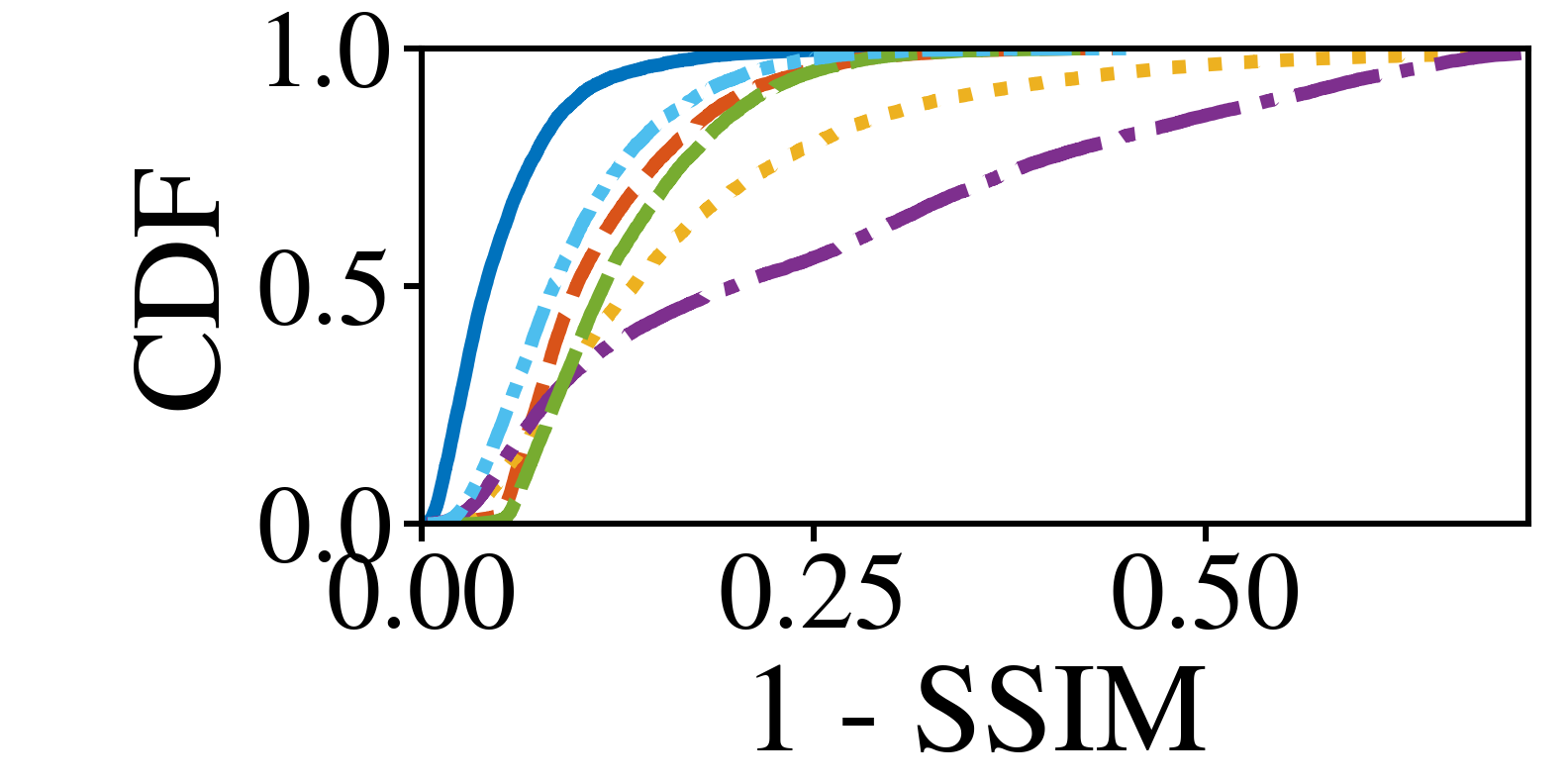} &
        \includegraphics[width=0.248\textwidth, valign=m]{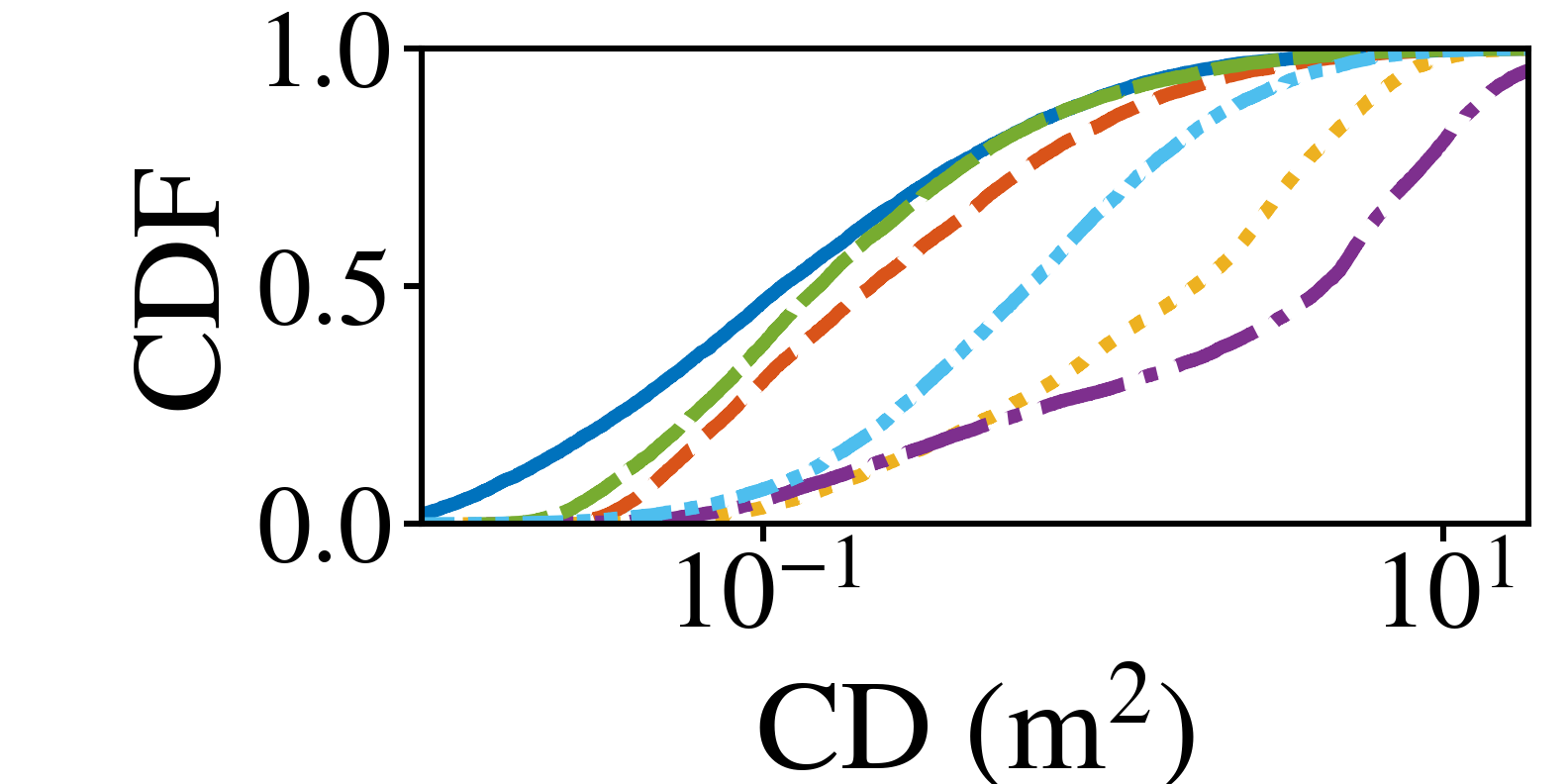} &
        \includegraphics[width=0.248\textwidth, valign=m]{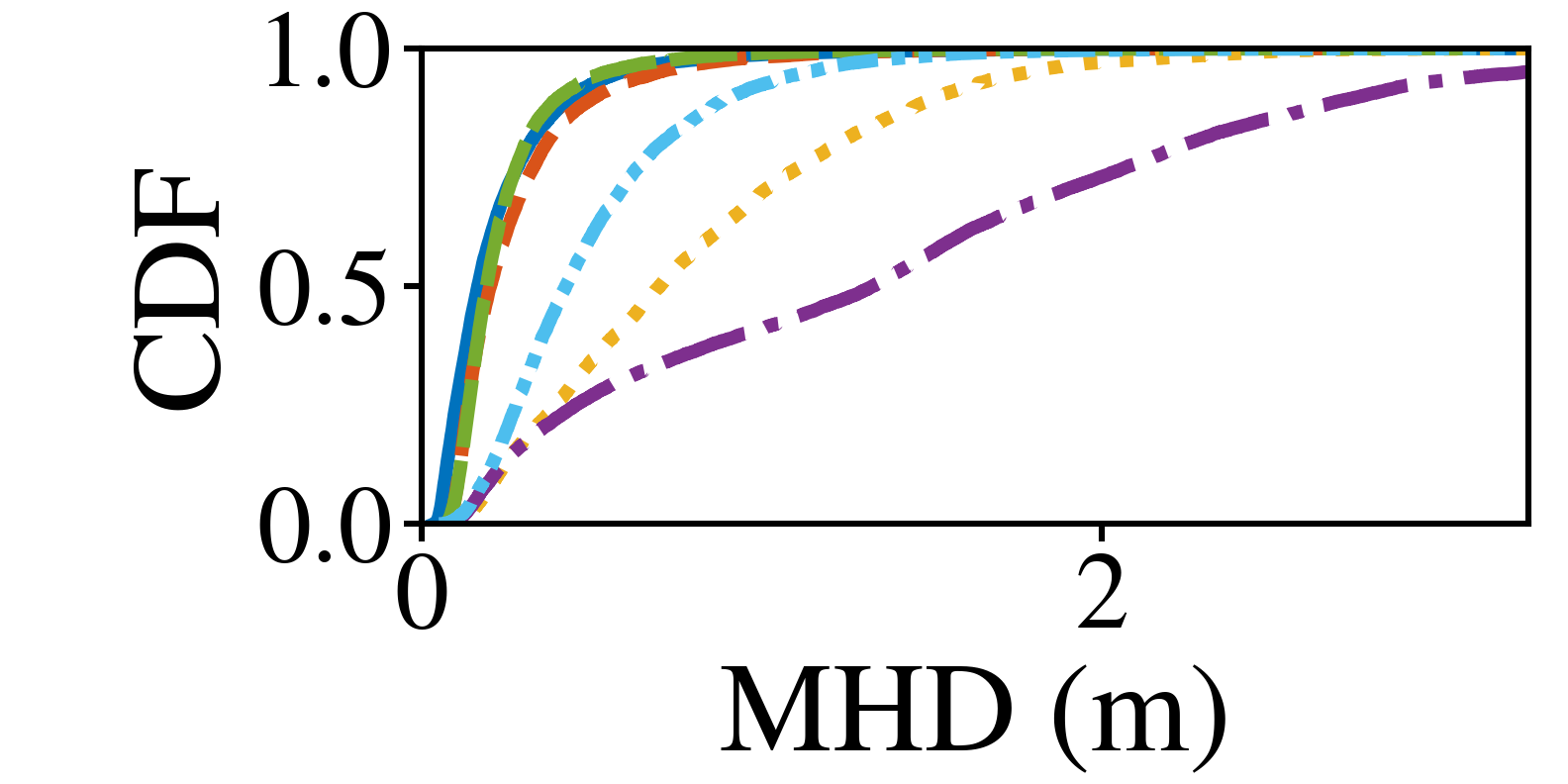} \\
    \end{tabular}
    \vspace{-0.05in}
    \caption{\revision{\small Empirical CDFs for clear (top) and smoke (bottom) scenes. \name{} shifts MAE, CD, and MHD toward lower values and SSIM toward higher values. DA3 and CaFNet develop the largest high-error tails under smoke.}}
    \vspace{-0.05in}
    \label{fig-cdf-new}
\end{figure*}

\noindent\textbf{Baselines.}
\revision{We compare with six camera-only, radar-only, and radar-camera methods. Each trainable baseline is trained or fine-tuned on the same building split; pretrained backbones retain their standard initialization. Before computing metrics, we resample every prediction to a common $288\times512$ depth grid.}

\begin{itemize}[leftmargin=*,nosep]
    \item \revision{\textbf{Depth Anything V3 (DA3)}~\cite{da3} is a general-purpose visual geometry model. We use its monocular prediction as the camera-only baseline.}

    \item \revision{\textbf{GRT}~\cite{grt} predicts 3D occupancy from a single raw radar frame and provides the radar-only baseline.}

    \item \revision{\textbf{GRT+Image} is our radar-camera extension of GRT~\cite{grt}. It concatenates ImageNet-pretrained ResNet-18 features with the radar representation before the GRT decoder, retaining raw radar I/Q input while adding RGB.}

    \item \revision{\textbf{CaFNet}~\cite{sun2024cafnet} fuses RGB with sparse radar point clouds using confidence-aware gated fusion.}

    \item \revision{\textbf{CaFNet (No-Smoke)} uses the same architecture but is trained only on clear images. Comparing the two CaFNet variants tests whether smoke exposure during training, rather than the fusion design, explains their performance.}

    \item \revision{\textbf{RadarCam-Depth}~\cite{li2024radarcam} estimates monocular relative depth and learns its global and local metric scale from sparse radar points.}
\end{itemize}

\noindent\textbf{Metrics.}
We evaluate depth quality in three domains: per-pixel metric accuracy, perceptual similarity, and 3D geometric fidelity. These metrics are commonly used in depth estimation and 3D reconstruction works~\cite{da3,grt,panoradar,sun2024cafnet}.

\begin{itemize}[leftmargin=*,nosep]
    \item \textbf{Per-pixel accuracy.} Mean Absolute Error (\textbf{MAE}), reported in metres, measures the average absolute difference between predicted and ground-truth depth values across all valid pixels.

    \item \textbf{Perceptual similarity.} Structural Similarity Index (\textbf{SSIM}) \cite{SSIM} evaluates contrast and structural consistency between depth maps (higher is better). Learned Perceptual Image Patch Similarity (\textbf{LPIPS}) \cite{lpips}, computed using AlexNet~\cite{alexnet} features, quantifies perceptual distance in a learned feature space (lower is better).

    \item \textbf{3D geometric fidelity.} We back-project predicted depth maps into 3D point clouds and compute Chamfer Distance (\textbf{CD}) \cite{cd}, the sum of the mean squared bidirectional nearest-neighbor distances, reported in m$^2$, and Modified Hausdorff Distance (\textbf{MHD}) \cite{mhd}, the larger mean directed nearest-neighbor distance, reported in metres.

    \item \revision{\textbf{Boundary sharpness.} Following DA3~\cite{da3}, Gradient Error (\textbf{GE}) is the mean $\mathcal{L}_1$ difference between horizontal and vertical finite differences of predicted and ground-truth depth after both are normalized to $[0,1]$. It captures edge error not explicitly represented by MAE.}

\end{itemize}

\vspace{-0.1in}
\subsection{Depth Prediction Quality}
\label{sec:e2e}

\revision{We first ask how well each method preserves metric and structural depth accuracy as RGB visibility degrades.}

\noindent\textbf{Quantitative Results.}
\revision{Table~\ref{tab:quan-new} reports median metrics for clear scenes and for medium-to-heavy smoke ($\mathrm{IR}\geq2000$). \name{} has the best performance on all five metrics in both conditions. In clear scenes, it reduces MAE from 0.415 to 0.303\,m and CD from 0.153 to 0.120\,m$^2$ relative to the strongest baseline on each metric. Under smoke, its MAE changes from 0.303 to 0.313\,m and its CD from 0.120 to 0.114\,m$^2$.}

\begin{table}[htb]
    \centering
    \footnotesize
    \setlength{\tabcolsep}{4pt}
    \begin{tabular}{lccccc}
        \toprule
        \textbf{Method} & \textbf{MAE}$\downarrow$ & \textbf{SSIM}$\uparrow$ & \textbf{LPIPS}$\downarrow$ & \textbf{CD}$\downarrow$ & \textbf{MHD}$\downarrow$ \\
        \midrule
        \multicolumn{6}{l}{\textit{Clear Scenario}} \\
        DA3~\cite{da3} & 0.500 & 0.932 & \cellcolor{blue!8}0.159 & 0.196 & 0.233 \\
        CaFNet (No-Smoke) & 0.421 & 0.944 & 0.174 & 0.171 & 0.211 \\
        CaFNet~\cite{sun2024cafnet} & 0.417 & \cellcolor{blue!8}0.945 & 0.160 & 0.174 & 0.210 \\
        GRT~\cite{grt} & 0.434 & 0.902 & 0.440 & 0.242 & 0.209 \\
        \revision{GRT+Image} & \cellcolor{blue!8}0.415 & 0.881 & 0.450 & \cellcolor{blue!8}0.153 & \cellcolor{blue!8}0.192 \\
        \revision{RadarCam-Depth} & 0.489 & 0.938 & 0.164 & 0.264 & 0.262 \\
        \revision{\name{}} & \cellcolor{blue!18}\textbf{0.303} & \cellcolor{blue!18}\textbf{0.960} & \cellcolor{blue!18}\textbf{0.126} & \cellcolor{blue!18}\textbf{0.120} & \cellcolor{blue!18}\textbf{0.164} \\
        \midrule
        \multicolumn{6}{l}{\textit{Smoke Scenario}} \\
        DA3~\cite{da3} & 1.255 & 0.801 & 0.307 & 4.407 & 1.325 \\
        CaFNet (No-Smoke) & 0.980 & 0.863 & 0.287 & 1.641 & 0.687 \\
        CaFNet~\cite{sun2024cafnet} & 0.949 & 0.865 & 0.282 & 1.846 & 0.700 \\
        GRT~\cite{grt} & 0.436 & 0.901 & 0.444 & 0.208 & 0.203 \\
        \revision{GRT+Image} & \cellcolor{blue!8}0.418 & 0.884 & 0.450 & \cellcolor{blue!8}0.146 & \cellcolor{blue!8}0.192 \\
        \revision{RadarCam-Depth} & 0.676 & \cellcolor{blue!8}0.916 & \cellcolor{blue!8}0.211 & 0.603 & 0.424 \\
        \revision{\name{}} & \cellcolor{blue!18}\textbf{0.313} & \cellcolor{blue!18}\textbf{0.959} & \cellcolor{blue!18}\textbf{0.137} & \cellcolor{blue!18}\textbf{0.114} & \cellcolor{blue!18}\textbf{0.167} \\
        \bottomrule
    \end{tabular}
    \caption{\revision{\small Median depth metrics on the building-disjoint test set. \name{} shows the best performance on every reported metric in both clear and smoke conditions.}}
    \vspace{-0.2in}
    \label{tab:quan-new}
\end{table}

\begin{figure*}[t]
    \centering
    \setlength{\tabcolsep}{0pt}
    \renewcommand{\arraystretch}{1.0}

    \newlength{\qualimgw}
    \setlength{\qualimgw}{0.106\textwidth}

    \newcommand{\qualimg}[1]{%
    \includegraphics[width=\qualimgw,valign=m]{#1}%
    }

    \newcommand{\colhead}[1]{%
    \makebox[\qualimgw][c]{\tiny\bfseries #1}%
    }

    \newcommand{\caselabel}[1]{%
    \rotatebox[origin=c]{90}{\footnotesize\bfseries #1}%
    }

    \begin{tabular}{@{}c@{\hspace{0.0035\textwidth}}c@{\hspace{0.0035\textwidth}}c@{\hspace{0.0035\textwidth}}c@{\hspace{0.0035\textwidth}}c@{\hspace{0.0035\textwidth}}c@{\hspace{0.0035\textwidth}}c@{\hspace{0.0035\textwidth}}c@{\hspace{0.0035\textwidth}}c@{\hspace{0.0035\textwidth}}c@{}}

        &
        \colhead{RGB Reference} &
        \colhead{Input RGB} &
        \colhead{Ground Truth} &
        \colhead{Ours} &
        \colhead{GRT} &
        \colhead{\revision{GRT+Image}} &
        \colhead{\revision{RadarCam-Depth}} &
        \colhead{CaFNet} &
        \colhead{DA3} \\[2pt]

        \caselabel{case 1} &
        \qualimg{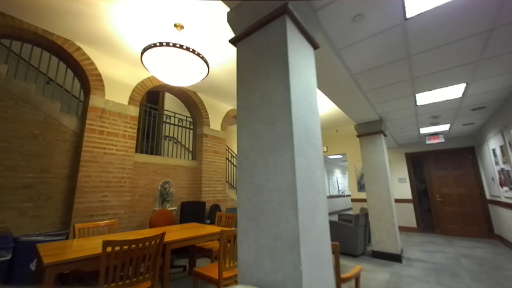} &
        \qualimg{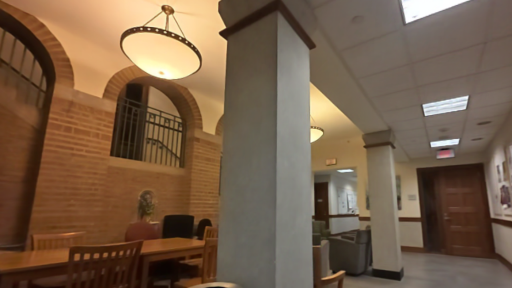} &
        \qualimg{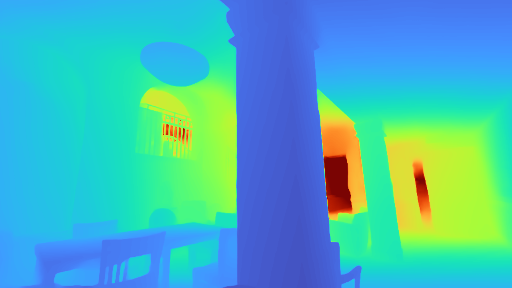} &
        \qualimg{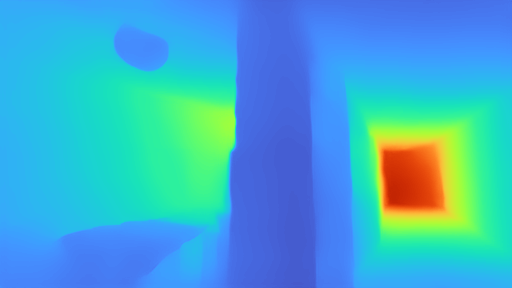} &
        \qualimg{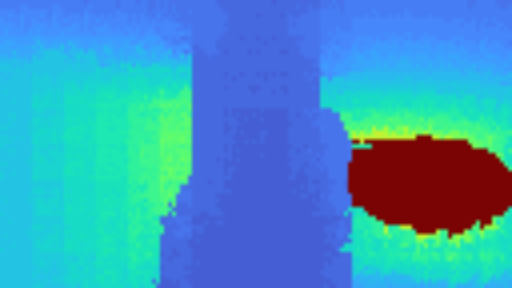} &
        \qualimg{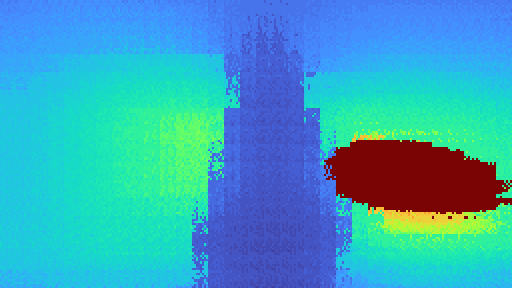} &
        \qualimg{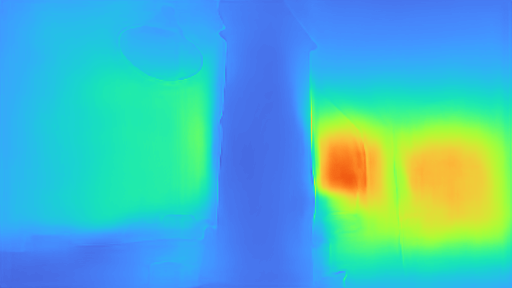} &
        \qualimg{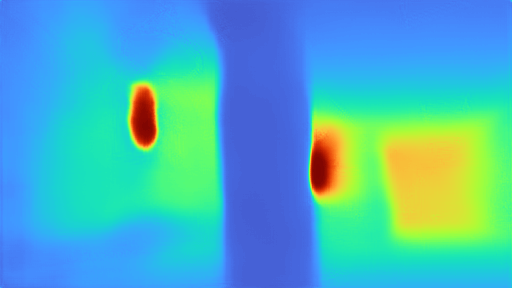} &
        \qualimg{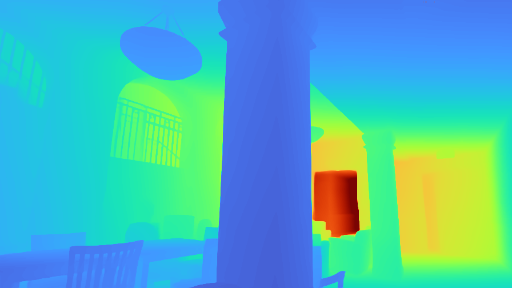} \\[1.8pt]

        \caselabel{case 2} &
        \qualimg{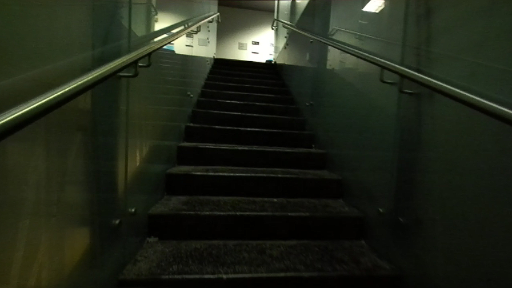} &
        \qualimg{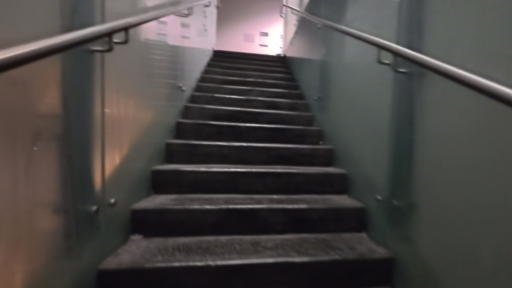} &
        \qualimg{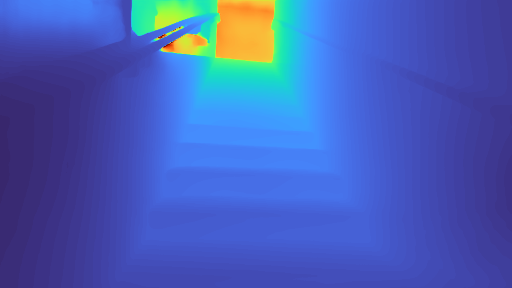} &
        \qualimg{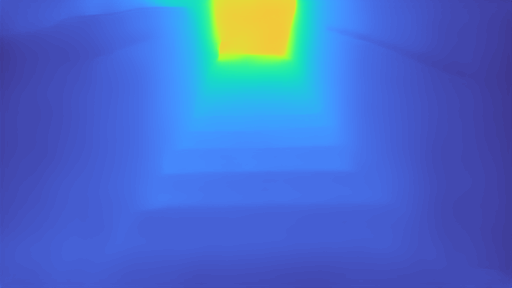} &
        \qualimg{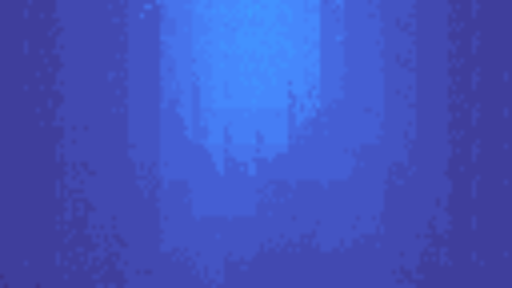} &
        \qualimg{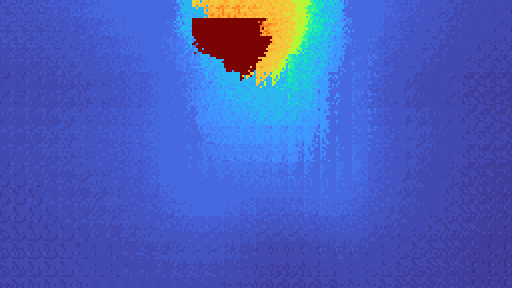} &
        \qualimg{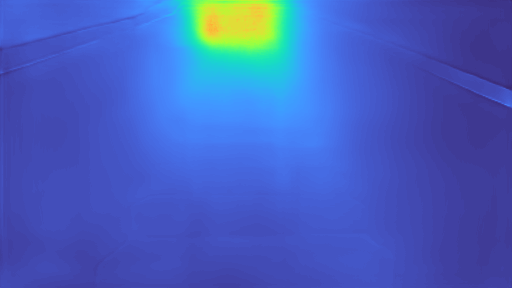} &
        \qualimg{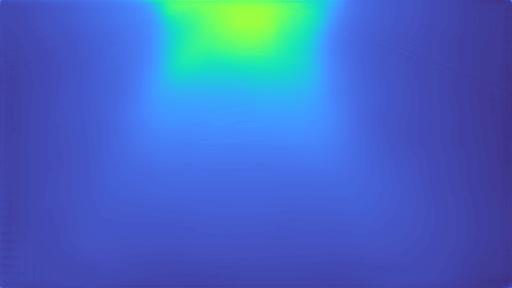} &
        \qualimg{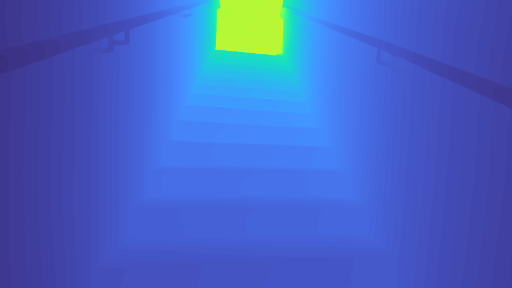} \\[1.8pt]

        \caselabel{case 3} &
        \qualimg{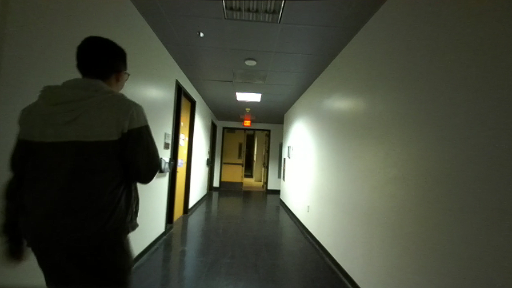} &
        \qualimg{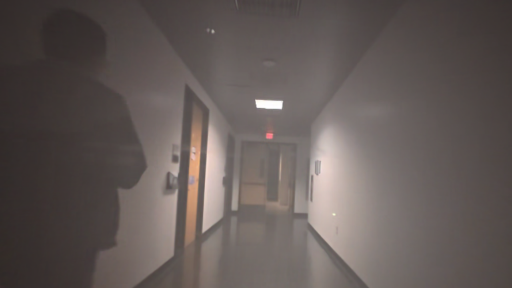} &
        \qualimg{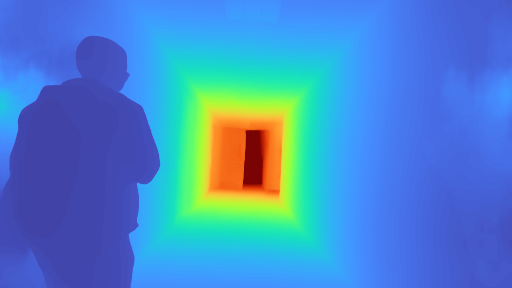} &
        \qualimg{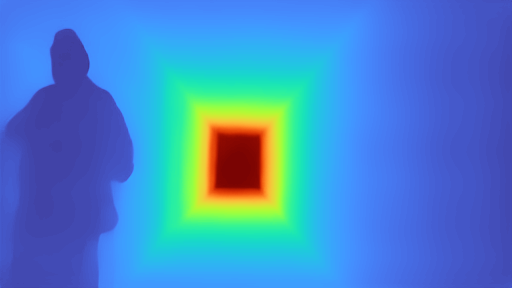} &
        \qualimg{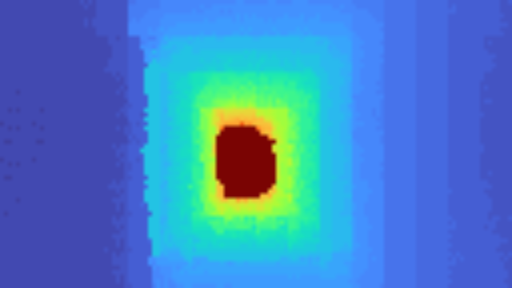} &
        \qualimg{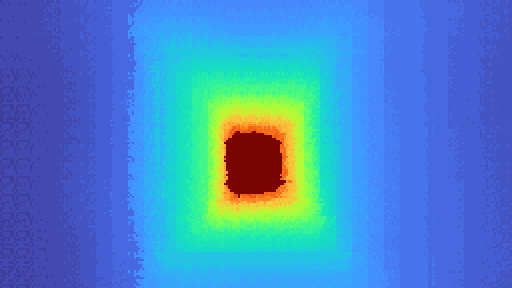} &
        \qualimg{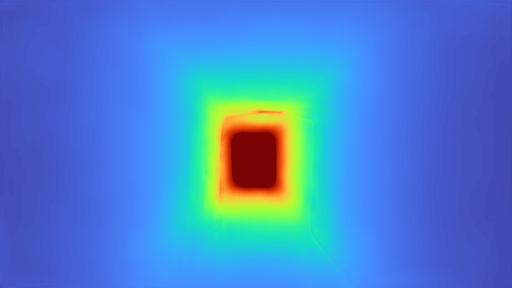} &
        \qualimg{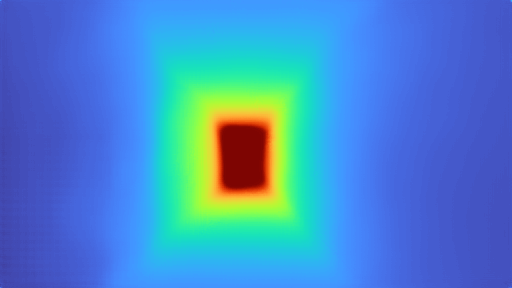} &
        \qualimg{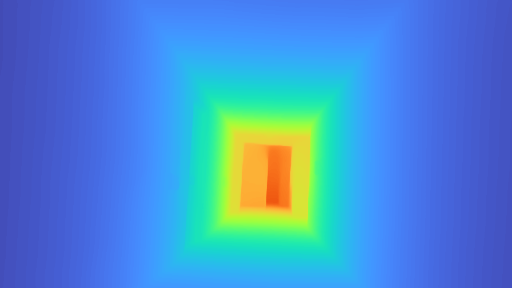} \\[1.8pt]

        \caselabel{case 4} &
        \qualimg{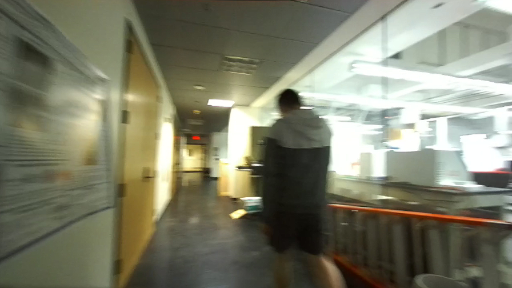} &
        \qualimg{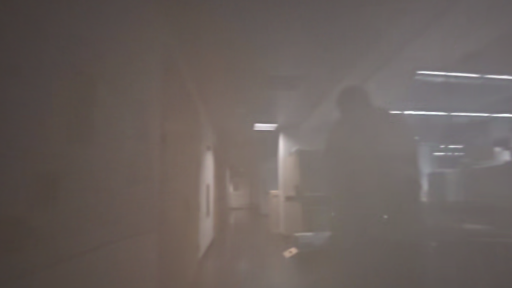} &
        \qualimg{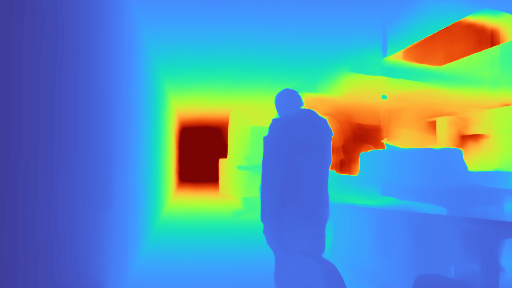} &
        \qualimg{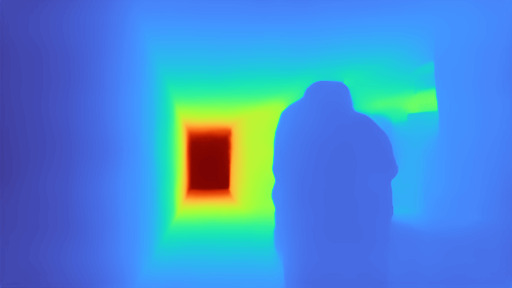} &
        \qualimg{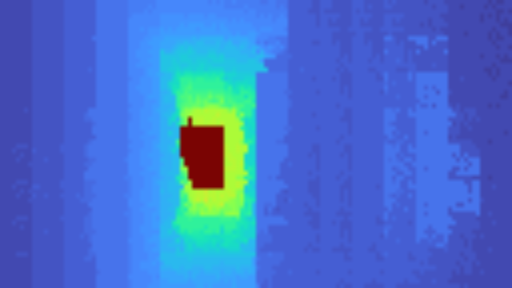} &
        \qualimg{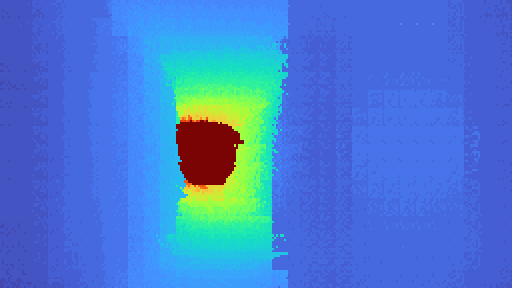} &
        \qualimg{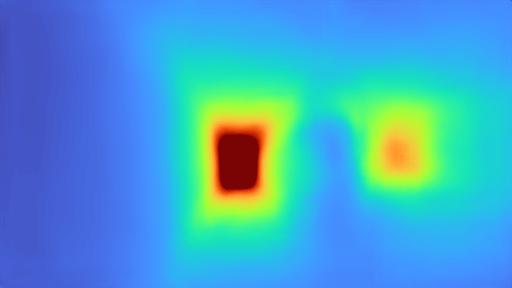} &
        \qualimg{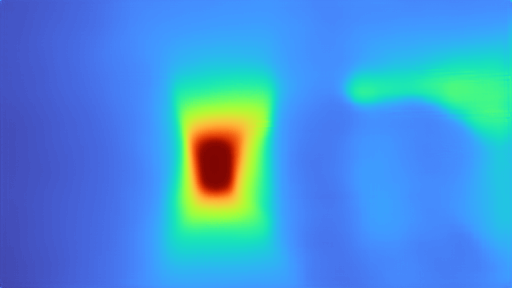} &
        \qualimg{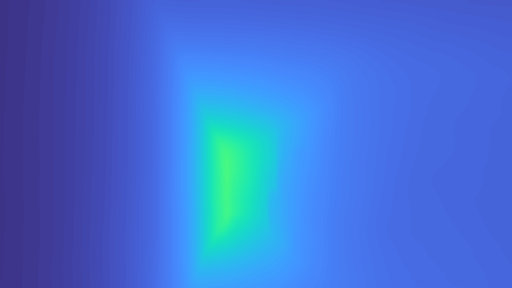} \\[1.8pt]

        \caselabel{case 5} &
        \qualimg{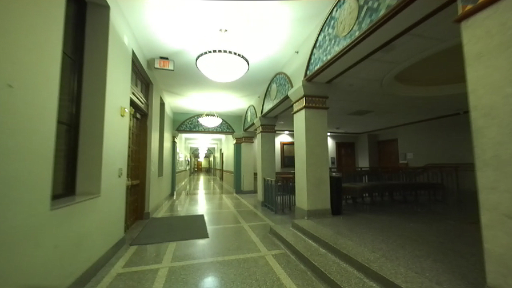} &
        \qualimg{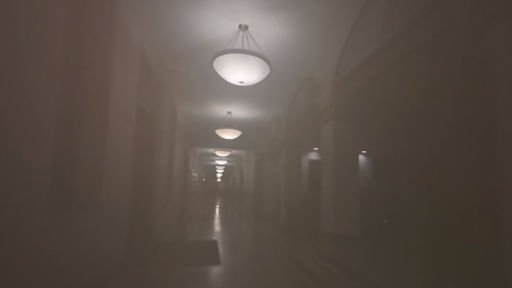} &
        \qualimg{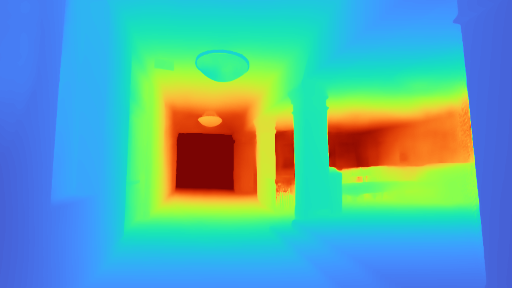} &
        \qualimg{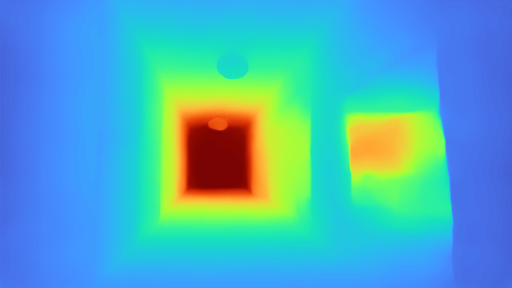} &
        \qualimg{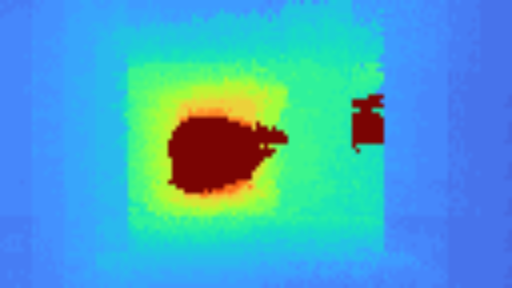} &
        \qualimg{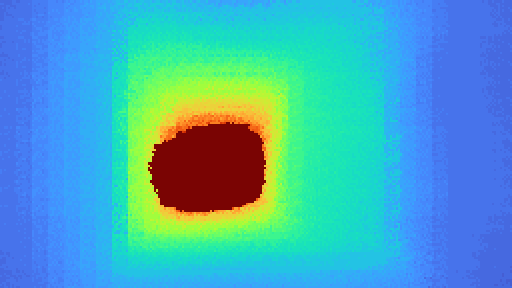} &
        \qualimg{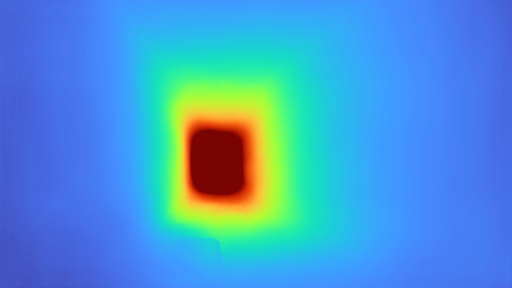} &
        \qualimg{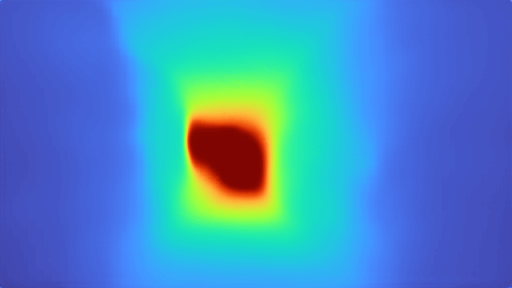} &
        \qualimg{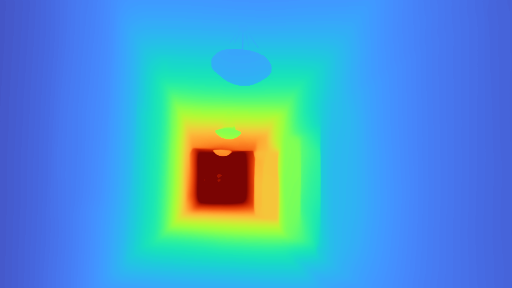} \\[1.8pt]

        \caselabel{case 6} &
        \qualimg{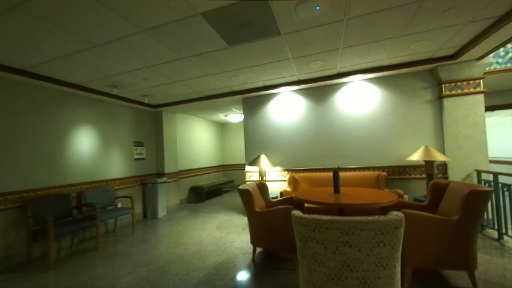} &
        \qualimg{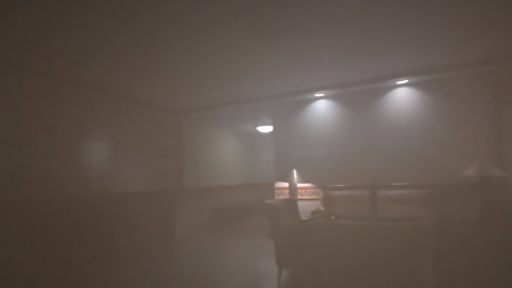} &
        \qualimg{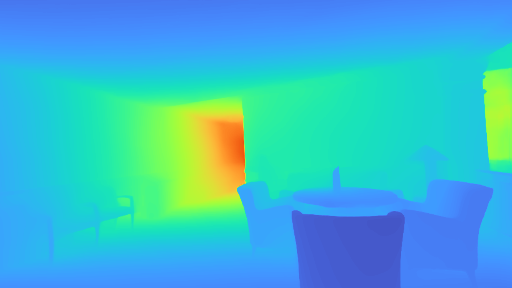} &
        \qualimg{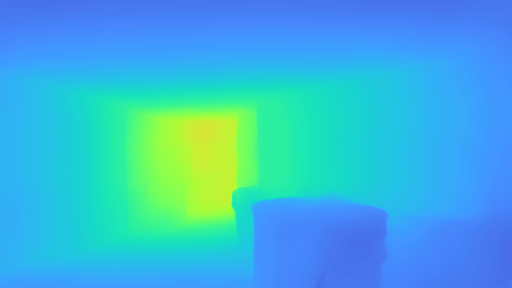} &
        \qualimg{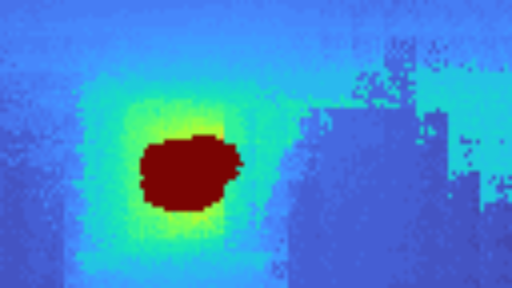} &
        \qualimg{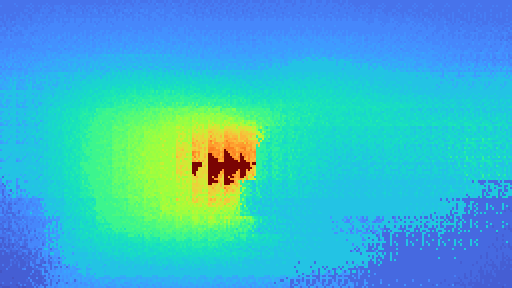} &
        \qualimg{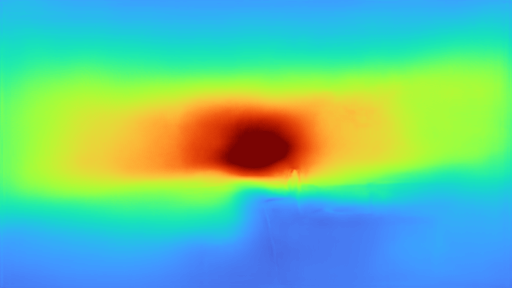} &
        \qualimg{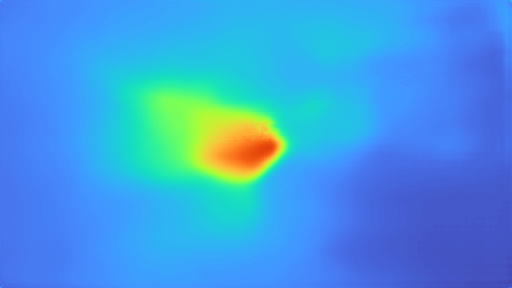} &
        \qualimg{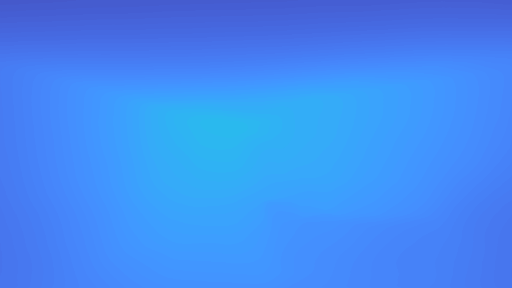} \\[1.8pt]

        \caselabel{case 7} &
        \qualimg{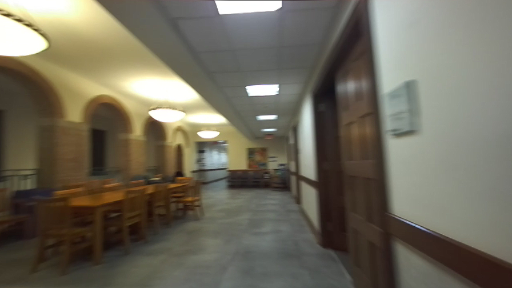} &
        \qualimg{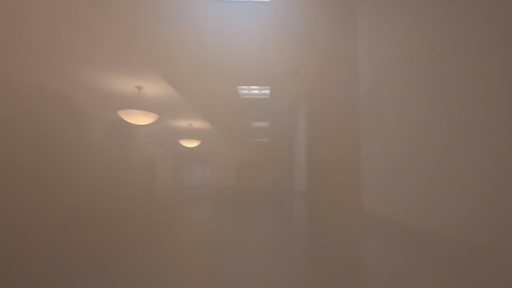} &
        \qualimg{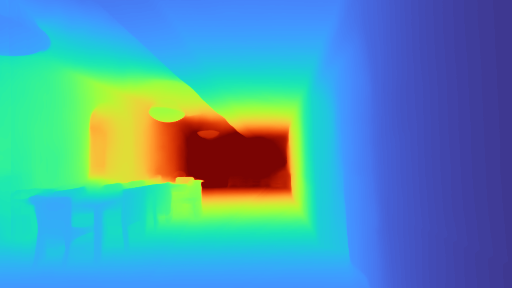} &
        \qualimg{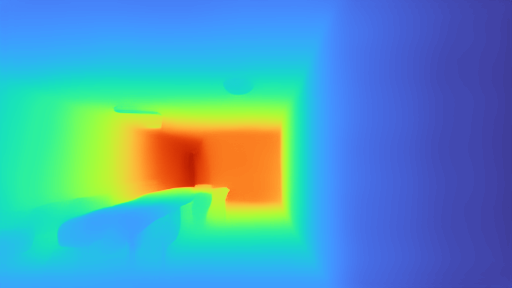} &
        \qualimg{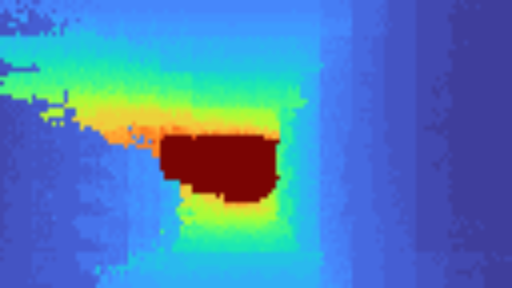} &
        \qualimg{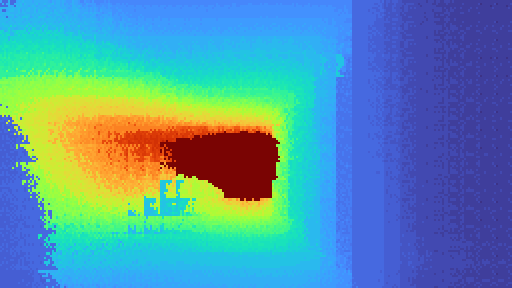} &
        \qualimg{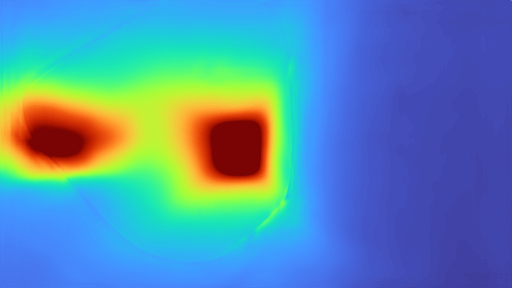} &
        \qualimg{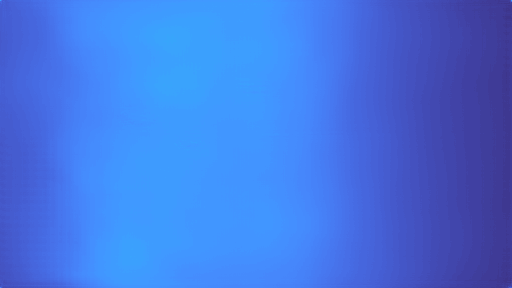} &
        \qualimg{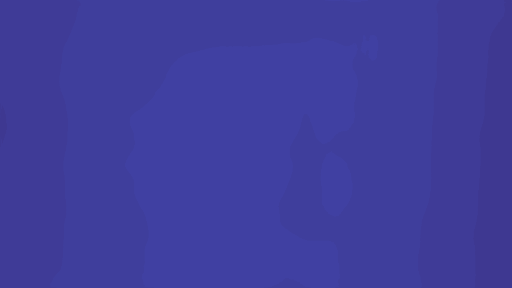} \\

    \end{tabular}

    \vspace{-0.05in}
    \caption{\small\revision{Representative depth predictions from clear through heavy-smoke scenes. \name{} retains finer boundaries than the radar baselines and avoids the severe smoke-induced loss of structure seen in the camera-dependent baselines.}}

    \vspace{-0.05in}
    \label{fig-qual-new}
\end{figure*}

\revision{The baselines show the contrasting roles of RGB and radar. DA3 and both CaFNet variants lose metric and geometric accuracy under smoke. Training CaFNet with smoke does not materially change this pattern: its smoke CD is 1.846\,m$^2$, compared with 1.641\,m$^2$ for the clear-only variant. Thus, smoke exposure alone does not close the gap for this fusion architecture. RadarCam-Depth degrades less severely, with CD increasing from 0.264 to 0.603\,m$^2$. This intermediate behavior is consistent with its design: radar supplies metric scale, while the dense prediction begins with monocular depth~\cite{li2024radarcam}. GRT remains nearly unchanged in MAE across visibility conditions, but its LPIPS stays above 0.44. Adding RGB to GRT lowers smoke CD from 0.208 to 0.146\,m$^2$ but leaves LPIPS at 0.450. \name{} combines the visibility-stable metric accuracy of the radar baselines with lower LPIPS than every baseline in both conditions.}

\revision{Figure~\ref{fig-cdf-new} shows that the median results are not driven by a small subset of frames. Under smoke, DA3 and CaFNet develop long high-error tails in MAE, CD, and MHD. RadarCam-Depth has a smaller tail, whereas the radar-based curves change less between conditions. \name{} remains shifted toward lower error and higher SSIM across the distributions.}

\noindent\textbf{Qualitative Results.}
\revision{Figure~\ref{fig-qual-new} shows representative predictions from clear through heavy-smoke scenes. DA3 and CaFNet retain more structure in the clear examples than in the smoke examples, while GRT remains coarse in both. GRT+Image and RadarCam-Depth avoid the near-empty heavy-smoke outputs seen from DA3 but lose boundary detail. Across these examples, \name{} most consistently preserves the major surfaces, people, stairs, and furniture boundaries visible in the reference depth.}

\subsection{Stability Under Varying Smoke Density}
\label{sec:smoke_stress}

\begin{table}[htb]
    \vspace{-0.05in}
    \centering
    \footnotesize
    \setlength{\tabcolsep}{4pt}
    \begin{tabular}{lccccc}
        \toprule
        \textbf{Method} & \textbf{MAE}$\downarrow$ & \textbf{SSIM}$\uparrow$ & \textbf{LPIPS}$\downarrow$ & \textbf{CD}$\downarrow$ & \textbf{MHD}$\downarrow$ \\
        \midrule
        \multicolumn{6}{l}{\textit{Light smoke}} \\
        DA3~\cite{da3} & 0.548 & 0.932 & \cellcolor{blue!8}0.165 & 0.290 & 0.295 \\
        CaFNet (No-Smoke) & 0.549 & 0.929 & 0.190 & 0.311 & 0.293 \\
        CaFNet~\cite{sun2024cafnet} & 0.508 & 0.935 & 0.181 & 0.278 & 0.277 \\
        GRT~\cite{grt} & 0.442 & 0.899 & 0.449 & 0.233 & 0.211 \\
        \revision{GRT+Image} & \cellcolor{blue!8}0.415 & 0.882 & 0.453 & \cellcolor{blue!8}0.152 & \cellcolor{blue!8}0.196 \\
        \revision{RadarCam-Depth} & 0.513 & \cellcolor{blue!8}0.937 & 0.171 & 0.332 & 0.303 \\
        \revision{\name{}} & \cellcolor{blue!18}\textbf{0.295} & \cellcolor{blue!18}\textbf{0.962} & \cellcolor{blue!18}\textbf{0.122} & \cellcolor{blue!18}\textbf{0.107} & \cellcolor{blue!18}\textbf{0.160} \\
        \midrule
        \multicolumn{6}{l}{\textit{Medium smoke}} \\
        DA3~\cite{da3} & 0.857 & 0.890 & 0.256 & 1.837 & 0.735 \\
        CaFNet (No-Smoke) & 0.876 & 0.882 & 0.272 & 0.929 & 0.529 \\
        CaFNet~\cite{sun2024cafnet} & 0.810 & 0.891 & 0.262 & 0.987 & 0.532 \\
        GRT~\cite{grt} & 0.455 & 0.897 & 0.448 & 0.230 & 0.212 \\
        \revision{GRT+Image} & \cellcolor{blue!8}0.424 & 0.880 & 0.457 & \cellcolor{blue!8}0.151 & \cellcolor{blue!8}0.195 \\
        \revision{RadarCam-Depth} & 0.649 & \cellcolor{blue!8}0.919 & \cellcolor{blue!8}0.211 & 0.545 & 0.402 \\
        \revision{\name{}} & \cellcolor{blue!18}\textbf{0.318} & \cellcolor{blue!18}\textbf{0.958} & \cellcolor{blue!18}\textbf{0.142} & \cellcolor{blue!18}\textbf{0.120} & \cellcolor{blue!18}\textbf{0.170} \\
        \midrule
        \multicolumn{6}{l}{\textit{Heavy smoke}} \\
        DA3~\cite{da3} & 1.751 & 0.625 & 0.385 & 7.796 & 1.970 \\
        CaFNet (No-Smoke) & 1.229 & 0.821 & 0.313 & 3.164 & 0.985 \\
        CaFNet~\cite{sun2024cafnet} & 1.292 & 0.794 & 0.322 & 3.503 & 1.053 \\
        GRT~\cite{grt} & \cellcolor{blue!8}0.404 & 0.907 & 0.437 & 0.173 & 0.189 \\
        \revision{GRT+Image} & \cellcolor{blue!8}0.404 & 0.889 & 0.442 & \cellcolor{blue!8}0.137 & \cellcolor{blue!8}0.187 \\
        \revision{RadarCam-Depth} & 0.726 & \cellcolor{blue!8}0.909 & \cellcolor{blue!8}0.210 & 0.703 & 0.468 \\
        \revision{\name{}} & \cellcolor{blue!18}\textbf{0.304} & \cellcolor{blue!18}\textbf{0.962} & \cellcolor{blue!18}\textbf{0.129} & \cellcolor{blue!18}\textbf{0.104} & \cellcolor{blue!18}\textbf{0.161} \\
        \bottomrule
    \end{tabular}
    \caption{\revision{\small Median metrics in the three sensor-defined smoke strata. \name{} is best on every metric in each stratum and changes little from light to heavy smoke.}}
    \vspace{-0.05in}
    \label{tab:smoke-density-new}
\end{table}

\begin{figure*}[htb]
    \vspace{-0.05in}
    \centering
    \setlength{\tabcolsep}{4.5pt}
    \renewcommand{\arraystretch}{1.0}
    \includegraphics[width=0.5\linewidth]{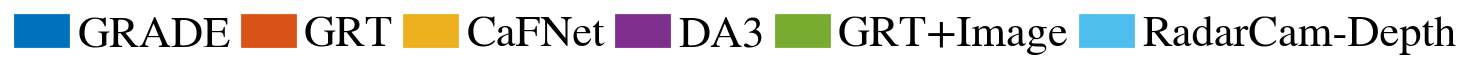}\\
    \begin{tabular}{@{}cccc@{}}
        \includegraphics[width=0.235\linewidth, valign=m]{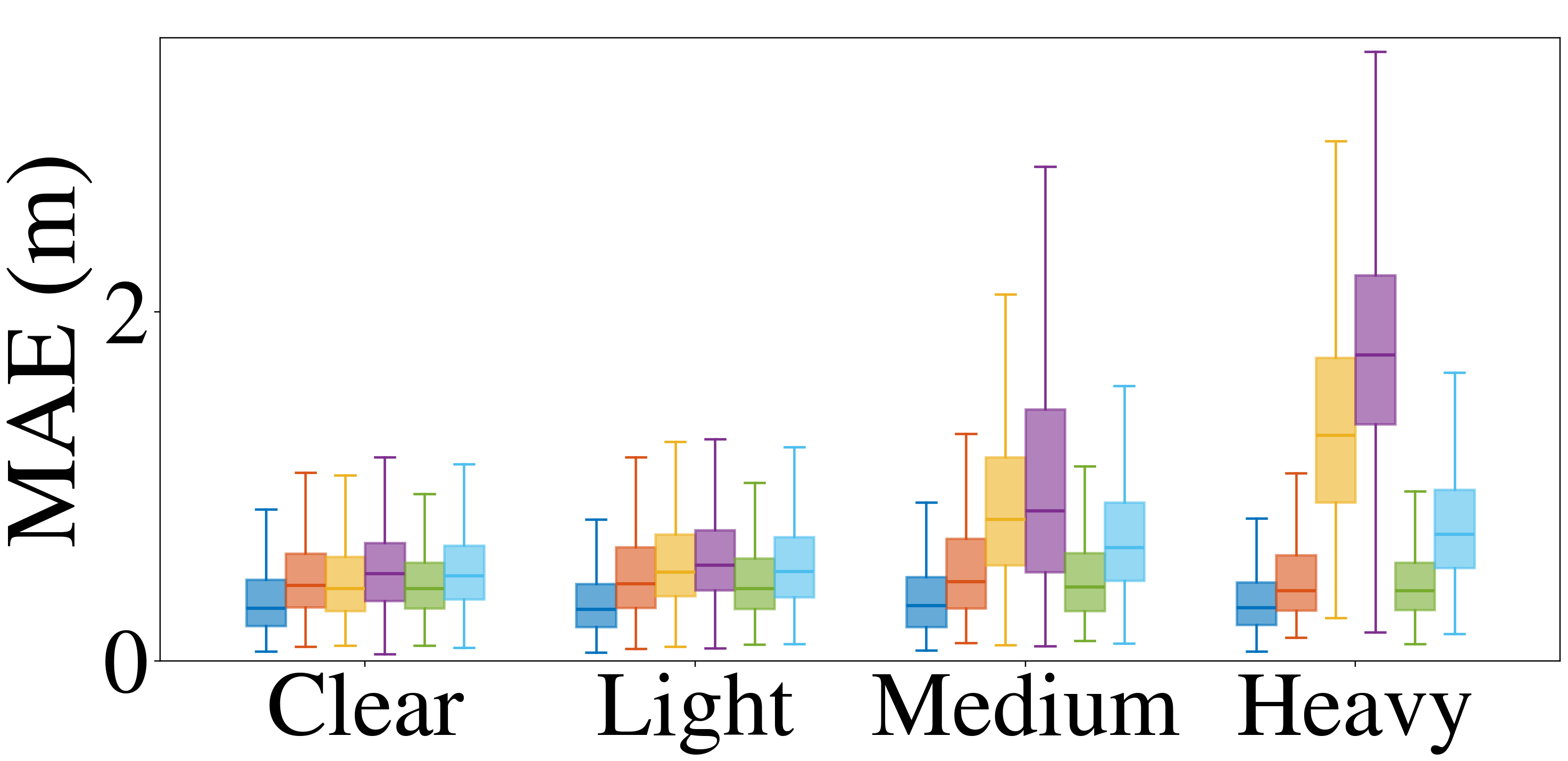} &
        \includegraphics[width=0.235\linewidth, valign=m]{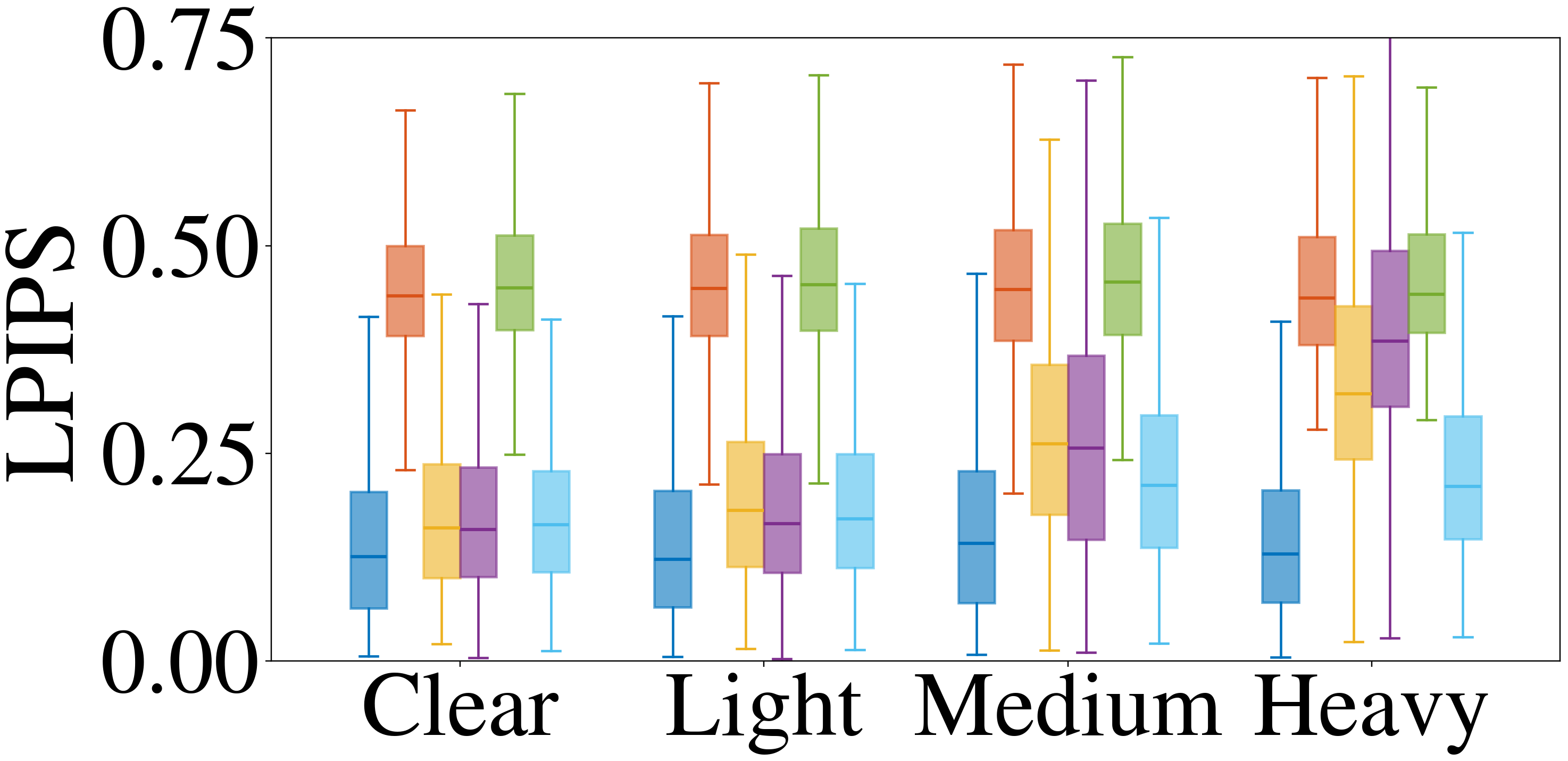} &
        \includegraphics[width=0.235\linewidth, valign=m]{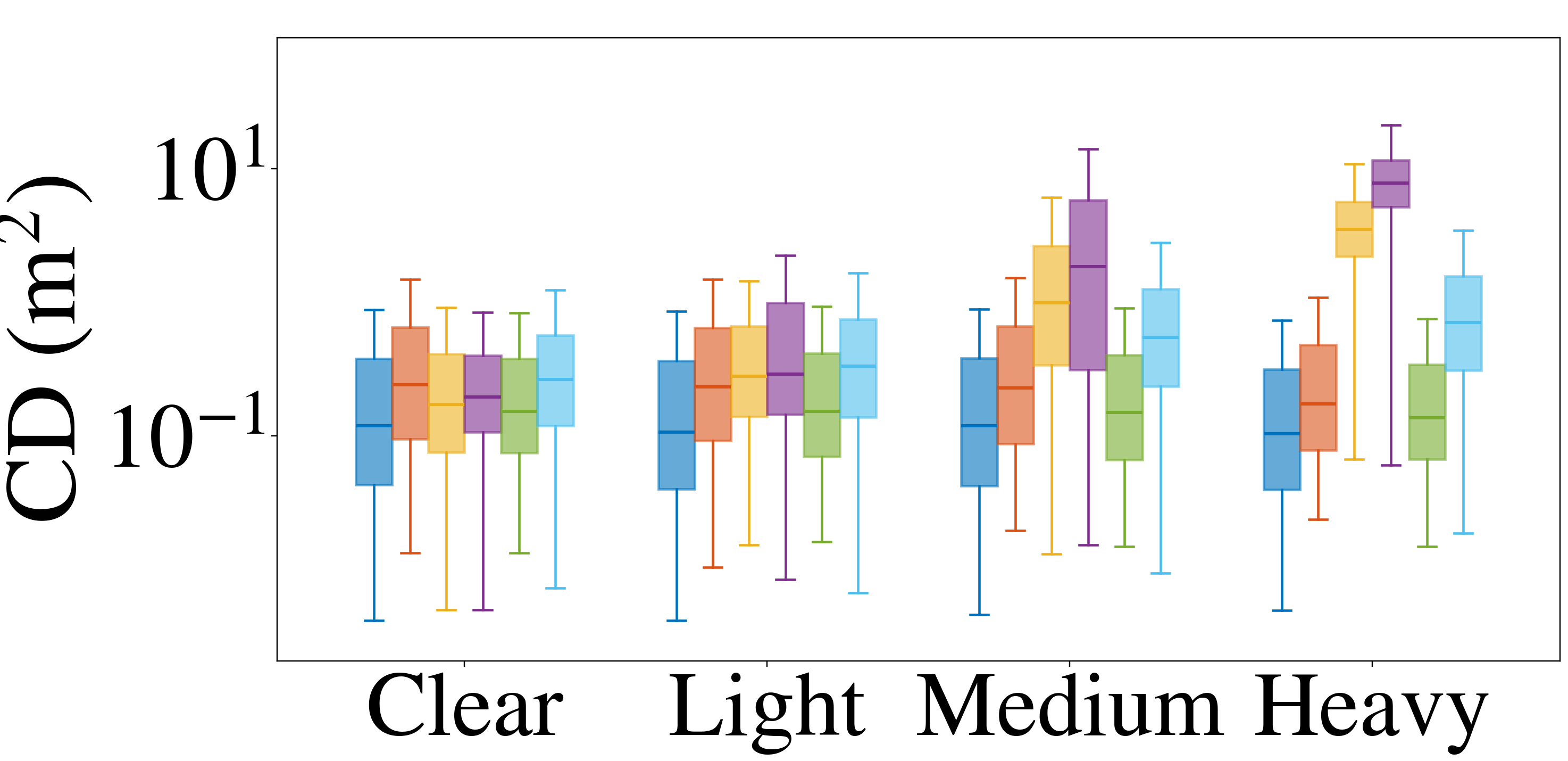} &
        \includegraphics[width=0.235\linewidth, valign=m]{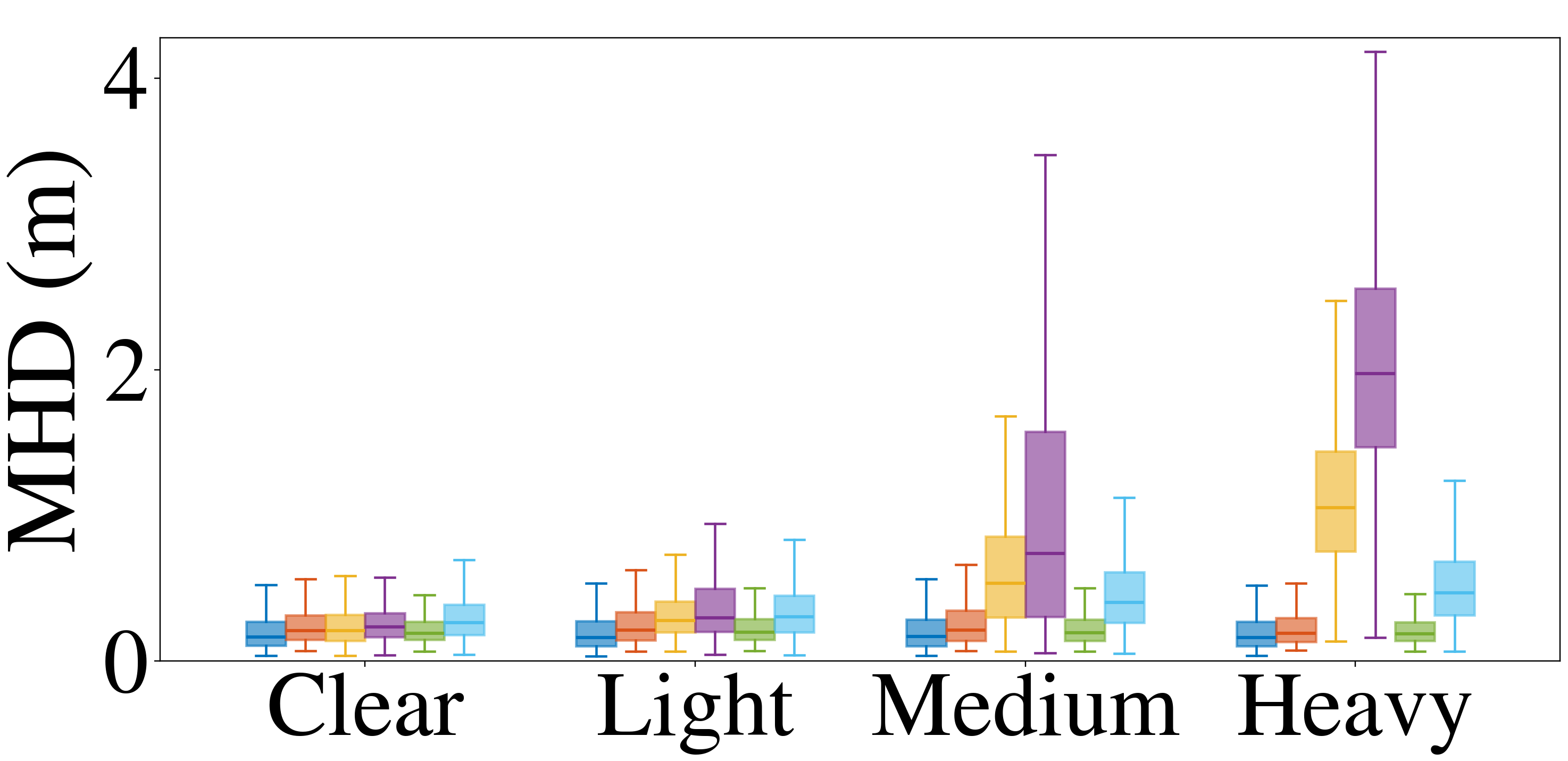} \\
    \end{tabular}
    \vspace{-0.05in}
    \caption{\revision{\small Performance from clear to heavy smoke with interquartile error bars. Camera baselines degrade sharply, while \name{} remains accurate and stable throughout.}}
    \vspace{-0.05in}
    \label{fig-smoke-vs-metric-new}
\end{figure*}

\revision{Table~\ref{tab:smoke-density-new} separates the smoke results by the sensor thresholds defined above. DA3's MAE rises from 0.548\,m in light smoke to 1.751\,m in heavy smoke, and its CD rises from 0.290 to 7.796\,m$^2$. CaFNet shows the same directional trend: its CD rises from 0.278 to 3.503\,m$^2$. The clear-only CaFNet variant reaches 3.164\,m$^2$ under heavy smoke, so smoke training does not provide a consistent advantage across these strata. RadarCam-Depth degrades more gradually, with CD increasing from 0.332 to 0.703\,m$^2$.}
\revision{The raw-radar methods change much less with smoke. GRT's MAE varies from 0.442 to 0.404\,m and GRT+Image from 0.415 to 0.404\,m. \name{} changes from 0.295 to 0.304\,m MAE and from 0.122 to 0.129 LPIPS between light and heavy smoke. In the heavy stratum, its CD is 0.104\,m$^2$, compared with 0.137\,m$^2$ for GRT+Image and 0.173\,m$^2$ for GRT.}

\revision{Figure~\ref{fig-smoke-vs-metric-new} adds the interquartile range within each stratum. The DA3 and CaFNet medians and intervals increase with the sensor reading, while the \name{} distributions remain similar across the three smoke levels.}

\subsection{Ablation Studies}
\label{sec:ablation}

We now isolate the contribution of each module and design choice in our framework.

\subsubsection{Contribution of Each Module}
\label{sec:ablation_stage}
\label{sec:graceful}

Table~\ref{tab:each-stage-new} isolates the radar prediction, diffusion refinement, and visual guidance stages over the pooled test set. \revision{Relative to Ours\_radar, Ours\_diffusion lowers GE from 0.0035 to 0.0027 but worsens MAE, LPIPS, CD, and MHD. A pretrained diffusion prior therefore does not improve the radar estimate by itself. Adding visual guidance produces Ours\_full, which is best on all six metrics, including MAE of 0.308\,m, LPIPS of 0.133, and CD of 0.116\,m$^2$.}

\begin{table}[htb]
    \vspace{-0.05in}
    \centering
    \scriptsize
    \setlength{\tabcolsep}{1pt}
    \begin{tabular*}{\columnwidth}{@{}l@{\extracolsep{\fill}}cccccc@{}}
        \toprule
        \textbf{Variant} & \makecell{\textbf{MAE}$\downarrow$} & \makecell{\textbf{SSIM}$\uparrow$} & \makecell{\textbf{LPIPS}$\downarrow$} & \makecell{\textbf{CD}$\downarrow$} & \makecell{\textbf{MHD}$\downarrow$} & \makecell{\textbf{GE}$\downarrow$} \\
        \midrule
        \revision{Ours\_radar} & \cellcolor{blue!8}0.320 & \cellcolor{blue!8}0.958 & \cellcolor{blue!8}0.147 & \cellcolor{blue!8}0.132 & \cellcolor{blue!8}0.172 & 0.0035 \\
        \revision{Ours\_diffusion} & 0.360 & 0.954 & 0.154 & 0.167 & 0.197 & \cellcolor{blue!8}0.0027 \\
        \revision{Ours\_full} & \cellcolor{blue!18}\textbf{0.308} & \cellcolor{blue!18}\textbf{0.959} & \cellcolor{blue!18}\textbf{0.133} & \cellcolor{blue!18}\textbf{0.116} & \cellcolor{blue!18}\textbf{0.166} & \cellcolor{blue!18}\textbf{0.0026} \\
        \revision{GRT\_Refine (Frozen)} & 0.450 & 0.939 & 0.182 & 0.230 & 0.243 & 0.0034 \\
        \revision{GRT\_Refine (Retrain)} & 0.433 & 0.946 & 0.173 & 0.248 & 0.244 & 0.0028 \\
        \bottomrule
    \end{tabular*}
    \caption{\revision{\small Stage-wise ablation over the pooled test set. Ours\_full is best on all metrics; without visual guidance, diffusion improves GE but worsens the other five metrics. Replacing Stage~1 with GRT remains worse after retraining the refinement stage.}}
    \vspace{-0.05in}
    \label{tab:each-stage-new}
\end{table}

\begin{figure}[t]
    \centering
    \captionsetup{font=small,skip=2pt}
    \includegraphics[width=0.75\linewidth]{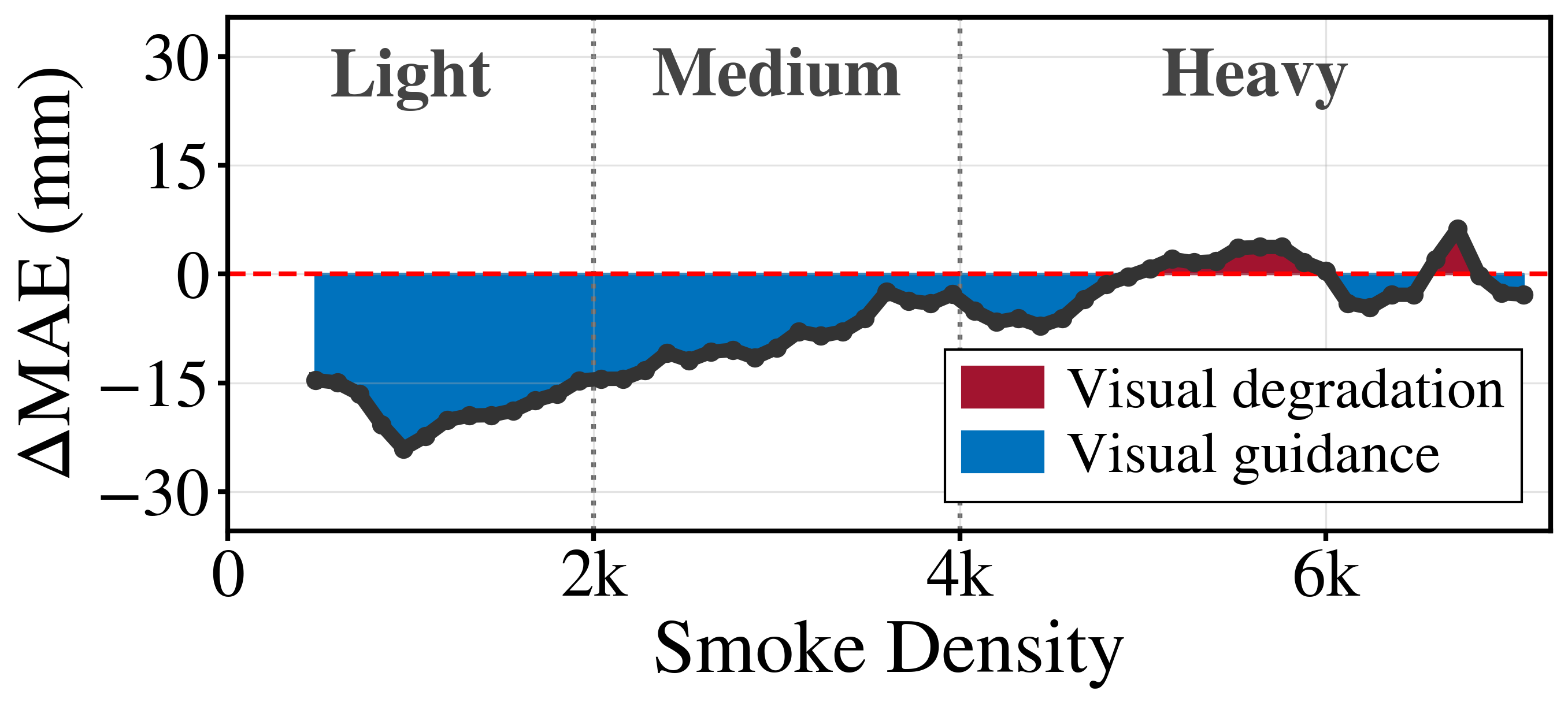}
    \vspace{-0.05in}
    \caption{\revision{\small $\Delta$MAE between \name{} and the radar-only module. Negative values favor \name{}.}}
    \vspace{-0.05in}
    \label{fig:graceful-degradation-updated}
\end{figure}

\revision{Figure~\ref{fig:graceful-degradation-updated} plots $\Delta\mathrm{MAE}=\mathrm{MAE}_{\textsc{grade}}-\mathrm{MAE}_{\text{radar}}$ against smoke density. Negative values through light and medium smoke show that \name{} effectively uses the available visual cues. As visibility degrades, $\Delta\mathrm{MAE}$ approaches and remains near zero rather than increasing steadily, with only small crossings in either direction. Thus, the full model gains from RGB when it is informative and approaches the radar-only estimate when it is not.}

\revision{Replacing Stage~1 with GRT while freezing the refinement stage gives a pooled MAE of 0.450\,m. Retraining the refinement stage on GRT outputs lowers it to 0.433\,m, with LPIPS of 0.173. The retrained variant still trails Ours\_full on every reported metric. The Stage~1 gain therefore remains after the downstream refinement is adapted to GRT's output distribution.}

\begin{table}[htb]
    \centering
    \renewcommand{\arraystretch}{0.96}
    \begin{minipage}[t]{0.46\columnwidth}
        \vspace{0pt}
        \centering
        \footnotesize
        \setlength{\tabcolsep}{2pt}
        \begin{tabular}{@{}l>{\centering\arraybackslash}p{\mcolw}>{\centering\arraybackslash}p{\mcolw}@{}}
            \toprule
            \textbf{Variant} & \textbf{MAE}$\downarrow$ & \textbf{CD}$\downarrow$ \\
            \midrule
            \revision{Ours\_radar} & \cellcolor{blue!18}\textbf{0.320} & \cellcolor{blue!18}\textbf{0.132} \\
            \midrule
            \revision{\makecell[l]{Ours\_radar\\w/o Doppler}} & 0.471 & 0.318 \\
            \midrule
            \revision{GRT} & \cellcolor{blue!8}0.435 & \cellcolor{blue!8}0.220 \\
            \midrule
            \revision{\makecell[l]{GRT w/o\\ Doppler}} & 0.707 & 0.545 \\
            \bottomrule
        \end{tabular}
        \captionof{table}{\revision{\small Removing Doppler from radar spectrum input degrades accuracy.}}
        \label{tab:doppler-ablation}
    \end{minipage}\hfill
    \begin{minipage}[t]{0.5\columnwidth}
        \vspace{0pt}
        \centering
        \footnotesize
        \setlength{\tabcolsep}{2pt}
        \begin{tabular}{@{}>{\raggedright\arraybackslash}p{44pt}>{\centering\arraybackslash}p{\mcolw}>{\centering\arraybackslash}p{\mcolw}@{}}
            \toprule
            \textbf{Variant} & \textbf{MAE}$\downarrow$ & \textbf{CD}$\downarrow$ \\
            \midrule
            \multicolumn{3}{@{}l}{\textit{Base: Ours\_radar}} \\
            W/O $\mathcal{L}_{\text{grad}}$ & 0.385 & 0.203 \\
            W/ $\mathcal{L}_{\text{grad}}$ & \cellcolor{blue!18}\textbf{0.320} & \cellcolor{blue!18}\textbf{0.132} \\
            \midrule
            \multicolumn{3}{@{}l}{\textit{Base: Ours\_full}} \\
            W/O $\mathcal{L}_{\text{3D}}$ & 0.376 & 0.169 \\
            W/ $\mathcal{L}_{\text{3D}}$ & \cellcolor{blue!18}\textbf{0.308} & \cellcolor{blue!18}\textbf{0.116} \\
            \bottomrule
        \end{tabular}
        \captionof{table}{\revision{\small Edge-aware supervision by $\mathcal{L}_{\text{grad}}$ and geometric regularization by $\mathcal{L}_{\text{3D}}$ improve performance.}}
        \label{tab:loss-ablations}
    \end{minipage} 
    \vspace{-0.05in}
\end{table}

\subsubsection{Impact of Doppler.}
\label{sec:doppler_ablation}

\revision{We remove the Doppler axis in Ours\_radar and GRT by re-processing the raw spectra without it and retraining both models, keeping both models and every other setting unchanged. Both radar-only modules degrade substantially without Doppler. Pooled MAE rises by 47\% for Ours\_radar and 63\% for GRT, while CD more than doubles for both models. The same pattern appears in both radar-only architectures indicating that it reflects the importance of the information gain from the Doppler repetitions. The degradation is also consistently larger in the 3D metrics than in the per-pixel ones, which suggests Doppler contributes mainly by separating scatterers that share a range-azimuth-elevation cell rather than by improving overall depth calibration.}

\subsubsection{Impact of Gradient Loss on Radar Depth}
\label{sec:loss_ablation}

We isolate the effect of the gradient loss $\mathcal{L}_{\text{grad}}$ in the radar depth module (Table~\ref{tab:loss-ablations}). \revision{Removing $\mathcal{L}_{\text{grad}}$ raises pooled MAE from 0.320 to 0.385\,m and CD from 0.132 to 0.203\,m$^2$.} Standard $\mathcal{L}_1$ and perceptual supervision alone produce smooth, blurred depth that lacks the sharp structural transitions the downstream diffusion model requires for effective anchoring. Explicitly enforcing local depth gradients preserves edge-aware structure at surface boundaries, which is essential for the coarse depth to serve as a reliable conditioning signal.

\subsubsection{Impact of 3D Reconstruction Loss}
\label{sec:loss_3d_reconstruction}
We next study the effect of the 3D reconstruction loss $L_{3D}$ by retraining the model without this objective while keeping the radar depth module and training setup unchanged. The quantitative results are reported in Table~\ref{tab:loss-ablations} using pooled median metrics across clear and smoke frames.

\revision{Adding $L_{3D}$ improves both pooled metrics, reducing MAE from 0.376 to 0.308\,m and CD from 0.169 to 0.116\,m$^2$. These results show that 3D reconstruction supervision provides geometric regularization that image-space objectives alone do not.}

\subsubsection{Sampling Steps and Runtime}
\label{sec:sampling}

We study how the number of DDIM sampling steps affects prediction quality (Table~\ref{tab:ablation-step}). \revision{A single DDIM step is insufficient because it does not provide enough iterations to produce coherent depth. Increasing to 5 steps recovers most of the refinement benefit, and performance peaks at 8 steps. Beyond 10 steps, quality saturates, consistent with previously reported behavior of DDIM under over-sampling~\cite{ddim}. We therefore use 8 steps for all experiments.}

\begin{table}[htb]
    \centering
    \footnotesize
    \setlength{\tabcolsep}{2pt}
    \renewcommand{\arraystretch}{0.96}
    \begin{tabular*}{\columnwidth}{@{\extracolsep{\fill}}>{\centering\arraybackslash}p{24pt}*{5}{>{\centering\arraybackslash}p{\mcolw}}@{}}
        \toprule
        \textbf{Steps} & \textbf{MAE}$\downarrow$ & \textbf{SSIM}$\uparrow$ & \textbf{LPIPS}$\downarrow$ & \textbf{CD}$\downarrow$ & \textbf{MHD}$\downarrow$ \\
        \midrule
        1  & 2.850 & 0.124 & 0.884 & 7.045 & 1.907 \\
        5  & 0.434 & \cellcolor{blue!8}0.941 & \cellcolor{blue!8}0.188 & 0.293 & 0.253 \\
        8  & \cellcolor{blue!18}\textbf{0.423} & \cellcolor{blue!18}\textbf{0.943} & \cellcolor{blue!18}\textbf{0.186} & \cellcolor{blue!18}\textbf{0.280} & \cellcolor{blue!18}\textbf{0.248} \\
        10 & \cellcolor{blue!8}0.424 & \cellcolor{blue!18}\textbf{0.943} & \cellcolor{blue!18}\textbf{0.186} & \cellcolor{blue!8}0.284 & \cellcolor{blue!8}0.250 \\
        50 & 0.441 & 0.940 & \cellcolor{blue!8}0.188 & 0.310 & 0.258 \\
        \bottomrule
    \end{tabular*}
    \caption{\revision{\small Sampling step ablation. Quality saturates at 8 steps, with no further improvement beyond 10.}}
    \label{tab:ablation-step}
    \vspace{-0.1in}
\end{table}

\subsection{\revision{Performance Across Radar Sparsity and Range}}
\label{sec:radar_robustness}

\begin{figure}[h]
    \centering
    \setlength{\tabcolsep}{0pt}
    \captionsetup{font=small,skip=2pt}
    \includegraphics[height=10.8pt]{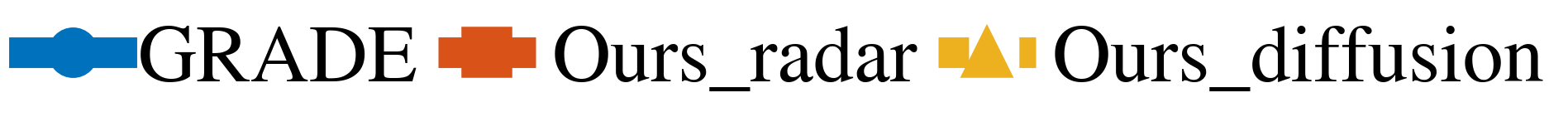}\\[1pt]
    \begin{tabular}{@{}cc@{}}
        \includegraphics[width=0.49\linewidth]{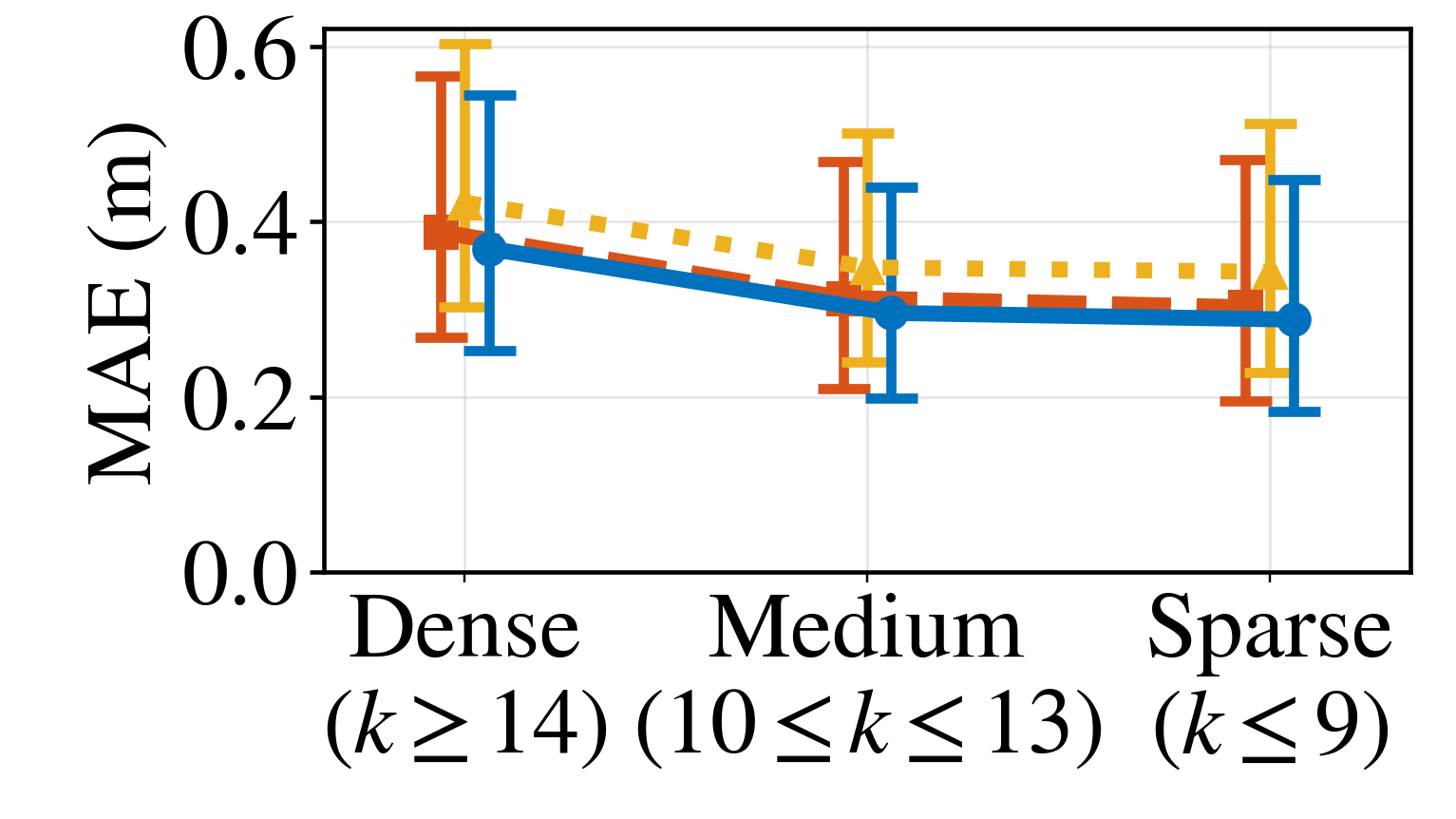} &
        \includegraphics[width=0.49\linewidth]{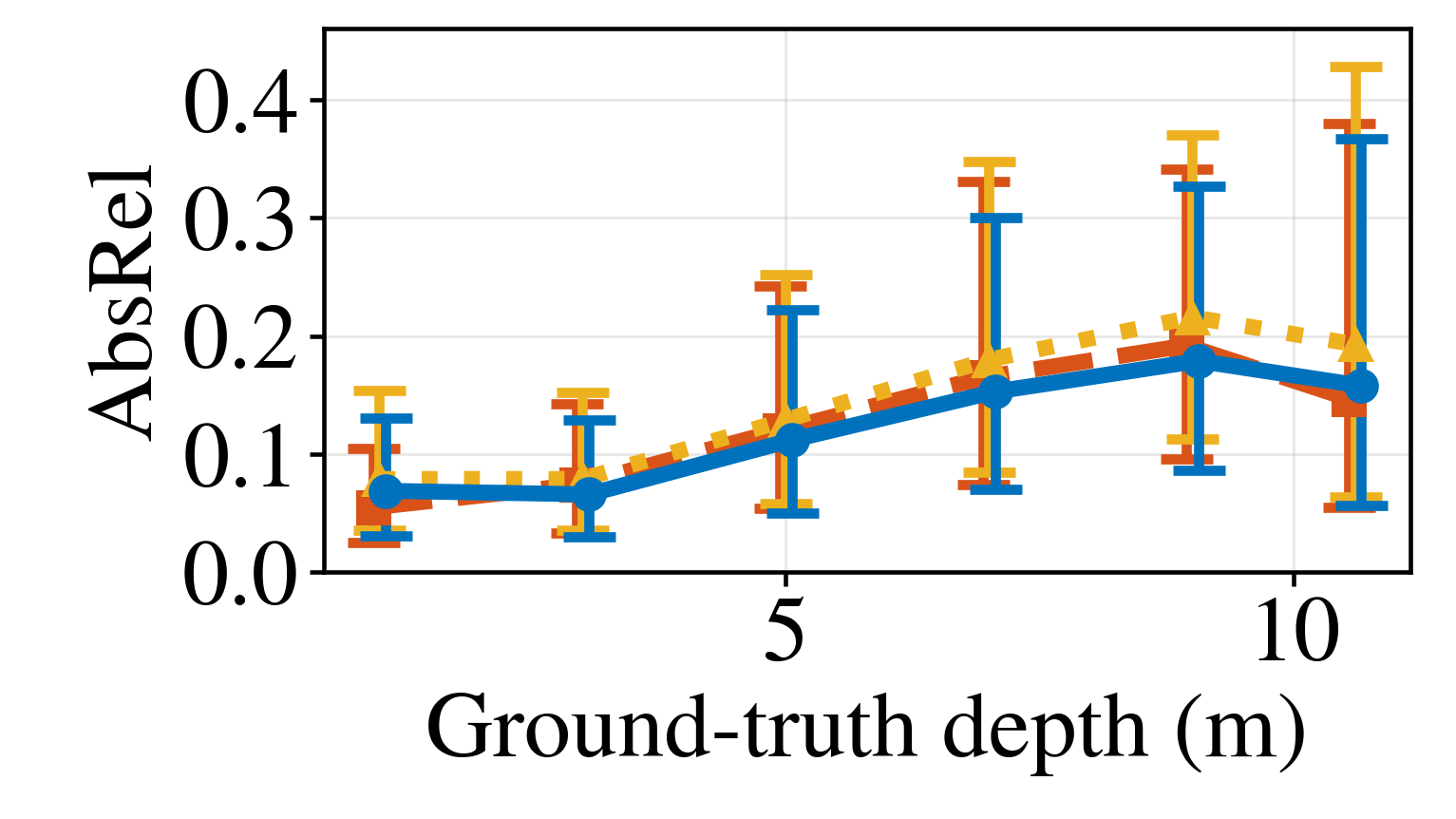}
    \end{tabular}
    \vspace{-0.05in}
    \caption{\revision{\small Stratified radar robustness study. Left: MAE across observed radar sparsity strata. Right: AbsRel across 2-meter range bins.}}
    \vspace{-0.15in}
    \label{fig:radar-robustness-updated}
\end{figure}

\revision{We evaluate robustness across radar return sparsity and ranges (Figure~\ref{fig:radar-robustness-updated}). We collect data across diverse crowded buildings with different materials and distances, and group frames by sparse ($k\leq9$), medium ($10\leq k\leq13$), and dense ($k\geq14$) radar returns. We also group corresponding ground-truth pixels into 2\,m range bins. The trained models, test frames, predictions, and metrics remain unchanged across all groups.}

\revision{Across the return-count strata, \name{} varies from 0.288\,m MAE in sparse scenes to 0.296\,m in medium scenes and 0.369\,m in dense scenes. Dense-return scenes contain more radar-visible surfaces and depth transitions, making the reconstruction more complex. Across range, we report absolute relative error (AbsRel), the per-pixel depth error divided by the ground-truth depth, which measures error relative to how far the surface is and is therefore comparable across bins. The trend is not uniform. The lower quartile stays below 0.1 in every bin for \name{} and Ours\_radar; Ours\_diffusion exceeds 0.1 only at 8--10\,m. For \name{}, the upper quartile grows steadily with range, from about 0.13 within 4\,m to 0.37 in the farthest bin, and the median rises from about 0.07 within 4\,m to 0.18 at 8--10\,m before easing slightly in the last bin. Across both analyses, error increases with return density and range, as one would expect with any radar system. Ours\_diffusion follows the same trend but is consistently worse, with MAE rising from 0.343\,m in sparse scenes to 0.422\,m in dense scenes and median AbsRel reaching 0.216 at 8--10\,m. Without RGB guidance, this additional diffusion error (0.216 versus 0.179 AbsRel for \name{}) indicates a greater risk of hallucinated geometry.}

\subsection{Performance Across Scene Complexity}
\label{sec:complexity}

We evaluate performance across scene complexity using grayscale RMS contrast of the reference RGB image as a measurable image-based proxy. \revision{We split the test set at its tercile boundaries into simple, medium, and complex groups and report medians for \name{} and all five baselines.}

\begin{figure}[h]
    \centering
    \setlength{\tabcolsep}{0pt}
    \captionsetup{font=small,skip=2pt}
    \includegraphics[height=10.8pt]{FIGURES/scene_complexity/scene_complexity_legend.png}\\[1pt]
    \begin{tabular}{@{}cc@{}}
        \includegraphics[width=0.49\linewidth]{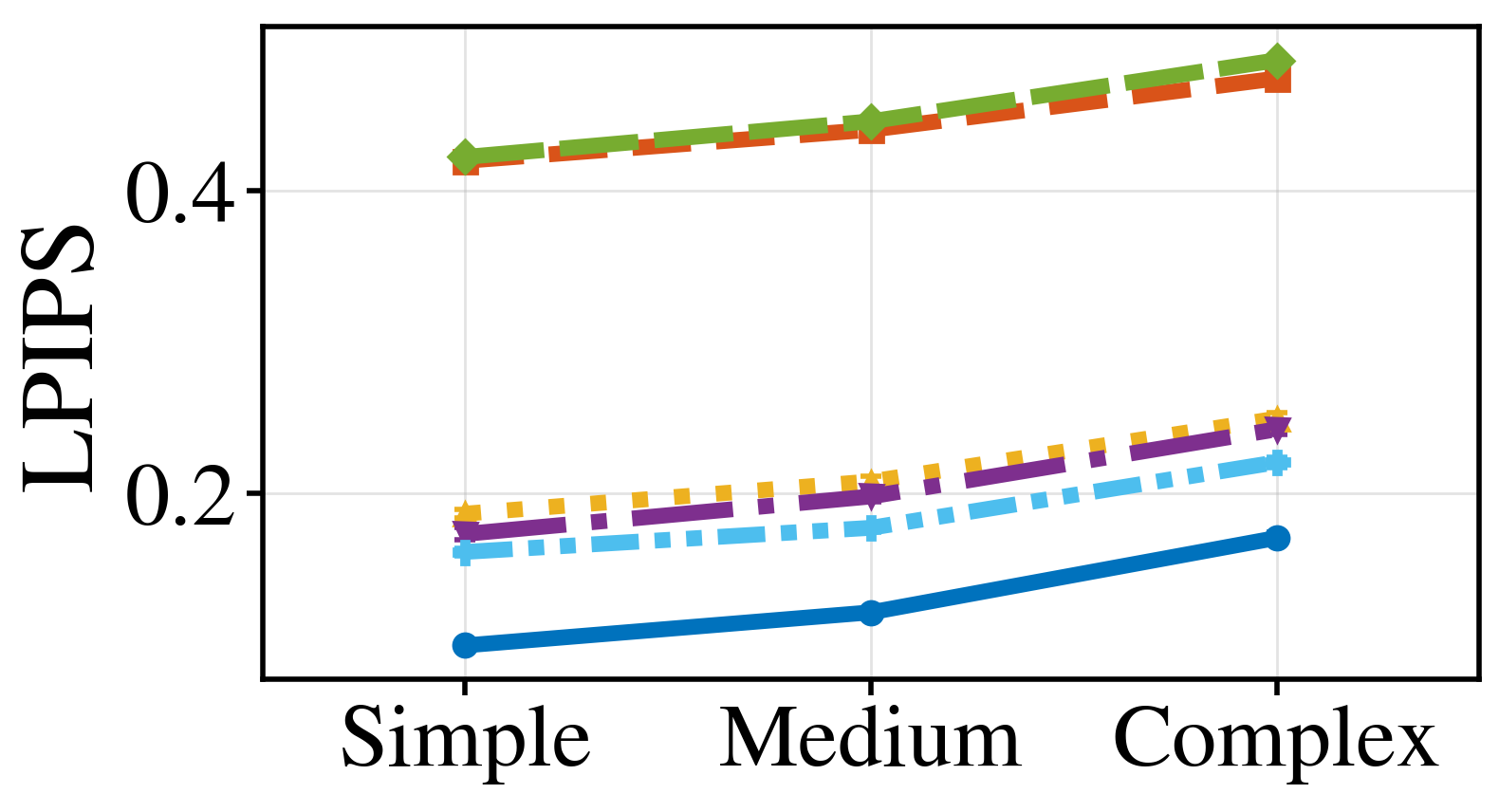} &
        \includegraphics[width=0.49\linewidth]{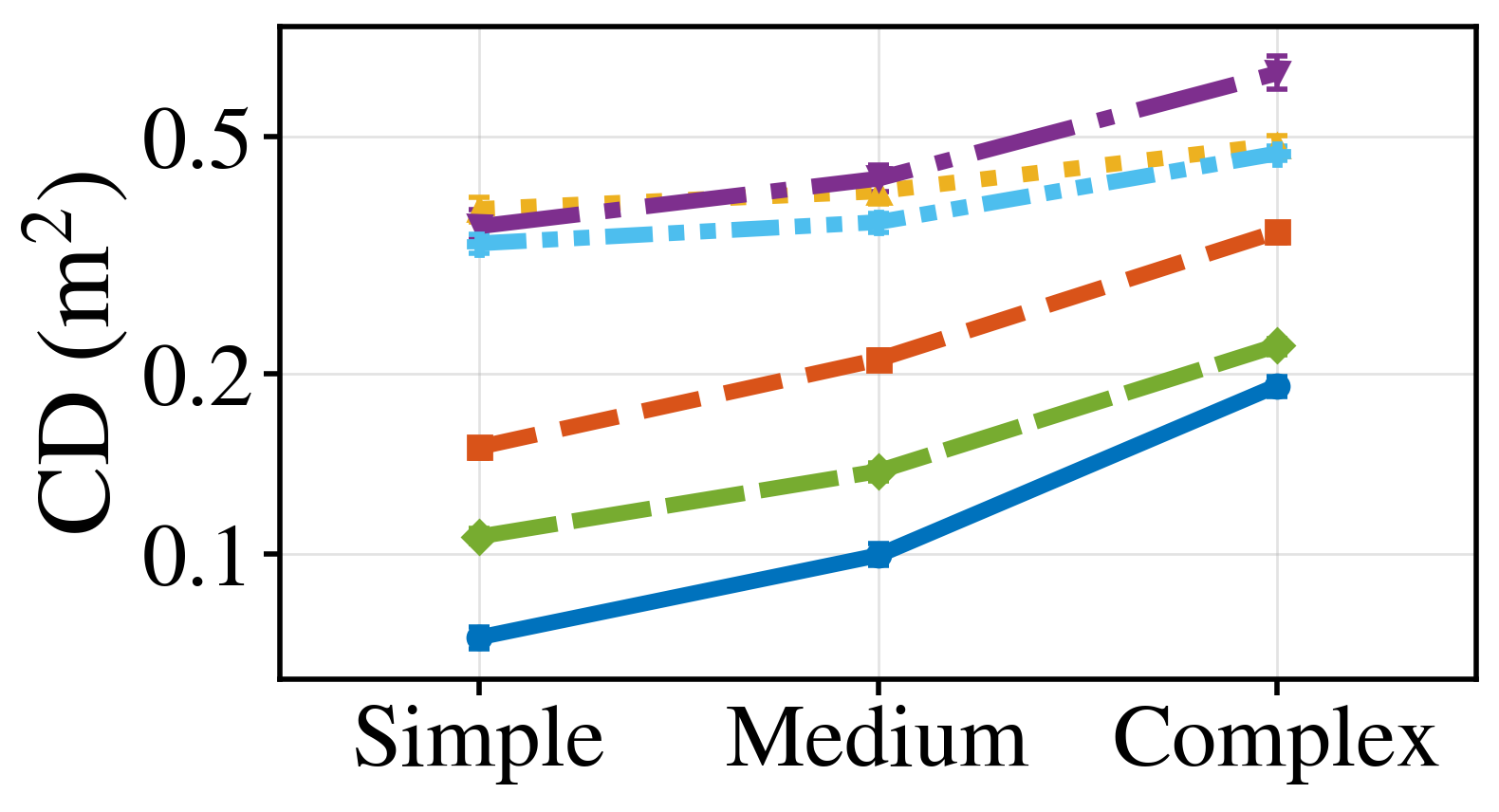}
    \end{tabular}
    \vspace{-0.05in}
    \caption{\revision{\small LPIPS (left) and CD (right) across scene-complexity terciles. As scenes get more complex, every model degrades.}}
    \vspace{-0.05in}
    \label{fig-scene-complexity}
\end{figure}

\revision{Figure~\ref{fig-scene-complexity} shows that LPIPS and CD increase with complexity for every method. \name{} has the lowest LPIPS and CD in all three groups. RadarCam-Depth is second-best in LPIPS, while GRT+Image is second-best in CD. The margin narrows in complex scenes, where the reference images contain more fine-scale structure than a single radar frame can resolve.}

\section{Discussion and Limitations}
\label{sec:discussion}

This work is a first realization of developing all-condition radar perception systems and there is plenty room for improvements and further work. We discuss a few points here.

\revision{\noindent\textbf{Computational cost.}
Iterative denoising is the main computational bottleneck. Radar-conditioned initialization reaches its best performance in 8 DDIM steps, but the current prototype is not designed for real-time wearable operation. Consistency distillation~\cite{song2023consistency} could reduce the number of passes, while quantization and smaller generative backbones could lower memory and per-step cost without changing the radar front end.}

\revision{\noindent\textbf{Temporal consistency and stability.}
\name{} estimates each frame independently. This provides immediate depth without waiting for controlled sensor motion, but it does not enforce consistency across a sequence. Continuous mapping and spatial overlays could incorporate pose-aware latent constraints or short-window radar aggregation while retaining the single-frame estimate when motion or odometry is unreliable.}

\revision{\noindent\textbf{Extreme and out-of-distribution conditions.}
Our building disjoint evaluations demonstrates transfer across unseen indoor spaces and real smoke, but does not cover outdoor geometry, unseen radar interference, different sensor configurations, or materials absent from the training set. Extending to these settings requires matched radar-depth data and explicit out-of-distribution testing. The radar module can be adapted to a new sensor or environment while preserving the generative refinement stage.}

\revision{\noindent\textbf{Generative hallucination risks.}
The pretrained prior recovers structure missing from a single radar frame, but some details are inferred rather than directly measured. Across our experiments, radar-grounded generative refinement improves structural fidelity and reduces depth error overall. As visual features weaken, its advantage drops, and on some subsets it slightly trails the radar-only prediction. Plausible detail should therefore not automatically be treated as measured geometry. Per-pixel uncertainty, radar-consistency tests during generation, and confidence-aware visual guidance could determine when to refine, return radar-only depth, or abstain. Our modular design also allows newer diffusion, flow-matching, or multimodal generative models to replace the current backbone.}

\vspace{-0.1in}

\section{Conclusion}
\label{sec:conclusion}

We presented \textit{GRADE}, which grounds a generative vision prior in single-frame radar geometry to estimate dense metric depth under visual degradation. Across 12 buildings with real smoke, \textit{GRADE} achieves the best result on every reported metric in both clear and smoke conditions.

\section*{Acknowledgements}
We thank the anonymous reviewers and our shepherd for their insightful comments and suggestions for improving this paper. 
This work was partially supported by NSF Vines Award 2549442.


\end{document}